\PassOptionsToPackage{dvipsnames,table}{xcolor}

\documentclass[acmtog,nonacm]{acmart}  %

\usepackage{booktabs} %

\usepackage[ruled]{algorithm2e} %

\SetAlFnt{\small}
\SetAlCapFnt{\small}
\SetAlCapNameFnt{\small}
\SetAlCapHSkip{0pt}

\acmJournal{TOG}
\acmYear{2024}
\acmVolume{43}
\acmNumber{4}
\acmArticle{63}
\acmMonth{7}

\setcopyright{rightsretained}

\acmDOI{10.1145/3658193}

\received{January 2024}
\received[final version]{April 2024}
\received[accepted]{April 2024}

\makeatletter
\@namedef{ver@everyshi.sty}{}
\makeatother

\usepackage{colortbl}
\usepackage{currfile}
\usepackage{etoolbox}
\usepackage{multirow}
\usepackage{tikz}
\usepackage{xfrac}

\usepackage[dvipsnames,table]{xcolor}

\definecolor{turquoise}{cmyk}{0.65,0,0.1,0.3}
\definecolor{purple}{rgb}{0.65,0,0.65}
\definecolor{dark_green}{rgb}{0, 0.5, 0}
\definecolor{orange}{rgb}{0.8, 0.6, 0.2}
\definecolor{red}{rgb}{0.8, 0.2, 0.2}
\definecolor{darkgray}{rgb}{0.5, 0.5, 0.5}
\definecolor{darkred}{rgb}{0.6, 0.1, 0.05}
\definecolor{blueish}{rgb}{0.0, 0.3, .6}
\definecolor{light_gray}{rgb}{0.7, 0.7, .7}
\definecolor{pink}{rgb}{1, 0, 1}
\definecolor{greyblue}{rgb}{0.15, 0.25, 0.65}

\newcommand{\TODO}[1]{\textbf{\color{red}[TODO: #1]}}

\newcommand{\moved}[1]{{\color{blue}#1}}

\newcommand{\changed}[1]{{\color{dark_green}#1}}

\newcommand{\copydanger}[1]{\textbf{OMITTED COPY/PASTE TEXT}}

\newcommand{\ignore}[1]{}

\usepackage{bbm}
\usepackage{lipsum}
\usepackage{blindtext}

\renewcommand{\paragraph}[1]{\vspace{.5em}\noindent\textbf{#1}.}

\newcommand{\NeRF}{NeRF\xspace}
\newcommand{\OurMethod}{SMERF\xspace}

\newcommand{\Ours}{Ours}
\newcommand{\MipNeRFThreeSixty}{mip-NeRF 360\xspace}
\newcommand{\ZipNeRF}{Zip-NeRF\xspace}
\newcommand{\BlockNeRF}{Block-NeRF\xspace}
\newcommand{\GSplat}{3DGS\xspace}

\newcommand{\MERF}{MERF\xspace}

\newcommand{\MERFDistill}{MERF+D\xspace}
\newcommand{\MERFDistillTrain}{MERF+DO\xspace}
\newcommand{\SNeRG}{SNeRG\xspace}
\newcommand{\BakedSDF}{BakedSDF\xspace}
\newcommand{\InstantNGP}{iNGP\xspace}

\newcommand{\SupplementaryMaterial}[1]{{supplementary material}}
\newcommand{\OurWebsite}{\url{https://smerf-3d.github.io}}

\usepackage{enumitem}
\setlist[itemize]{noitemsep,leftmargin=*,topsep=0em}
\setlist[enumerate]{noitemsep,leftmargin=*,topsep=0em}

\DeclareMathOperator*{\argmin}{arg\,min}

\newcommand{\density}{\sigma}
\newcommand{\dssim}{\text{DSSIM}}

\newcommand{\feature}{\mathbf{z}}

\newcommand{\lft}{\mathopen{}\left}
\newcommand{\lossmult}{\text{w}}

\newcommand{\norm}[1]{\left| \left| #1 \right| \right|}
\newcommand{\origin}{\mathbf{o}}

\newcommand{\pposition}{\mathbf{x}}
\newcommand{\radiance}{\mathbf{c}}
\newcommand{\ray}{\mathbf{r}}

\newcommand{\rgt}{\aftergroup\mathclose\aftergroup{\aftergroup}\right}
\newcommand{\rmse}{\text{RMSE}}
\newcommand{\ssim}{\text{SSIM}}
\newcommand{\submodel}{s}
\newcommand{\viewdir}{\mathbf{d}}

\newcommand{\CameraOrigin}{\origin}
\newcommand{\DeferredMlpParams}{\theta}

\newcommand{\DeferredNetworkRegularizer}{\mathrm{TV}}
\newcommand{\DeferredResolution}{\mathrm{P}}
\newcommand{\Density}{\sigma}
\newcommand{\DistanceToSubmodel}{\mathrm{D}}
\newcommand{\Exposure}{\epsilon}
\newcommand{\FeatureGatingWeight}{w}
\newcommand{\FocusRegion}{\mathbf{R}}
\newcommand{\HashGridIndex}{t}
\newcommand{\HashGridParameters}{\Phi}
\newcommand{\HashGridRegularizer}{\mathrm{HG}}
\newcommand{\ISO}{\mathrm{ISO}}
\newcommand{\Identity}{\mathbb{I}}
\newcommand{\Logit}{\operatorname{logit}}
\newcommand{\LossFn}{\mathcal{L}}

\newcommand{\Normal}{\mathcal{N}}

\newcommand{\NumHashGridEntries}{N}
\newcommand{\PPosition}{\pposition}
\newcommand{\PatchRadiance}{\mathbf{C}}
\newcommand{\PositionFeature}{\mathbf{t}}

\newcommand{\RayFeature}{\feature}
\newcommand{\Ray}{\ray}
\newcommand{\Reals}{\mathbb{R}}
\newcommand{\Real}{\mathbb{R}}
\newcommand{\SparseGridFeature}{\mathbf{V}}
\newcommand{\SparseGridResolution}{\mathrm{L}}
\newcommand{\SubmodelIndex}{k}
\newcommand{\SubmodelResolution}{\mathrm{K}}
\newcommand{\Submodel}{\mathcal{S}}
\newcommand{\TimesNoWS}{{\mkern-0mu\times\mkern-0mu}}

\newcommand{\Trilerp}{\operatorname{Trilerp}}
\newcommand{\TriplaneFeatureXY}{\mathbf{P}_{z}}
\newcommand{\TriplaneFeatureXZ}{\mathbf{P}_{y}}
\newcommand{\TriplaneFeatureYZ}{\mathbf{P}_{x}}
\newcommand{\TriplaneResolution}{\mathrm{R}}
\newcommand{\Uniform}{\mathcal{U}}
\newcommand{\VolumetricRender}{\operatorname{VolRend}}

\renewcommand{\density}{\tau}

\newcommand\commentout[1]{}

\definecolor{tabfirst}{rgb}{1, 0.7, 0.7} %
\definecolor{tabsecond}{rgb}{1, 0.85, 0.7} %
\definecolor{tabthird}{rgb}{1, 1, 0.7} %

\newcommand{\appendixref}[2]{\ifdef{\UseHardcodedReferences}{{Appendix~#1}}{Appendix~\ref{#2}}}
\newcommand{\appendixtableref}[2]{\ifdef{\UseHardcodedReferences}{{Table~#1 of the Appendices}}{Table~\ref{#2}}}

\newcommand{\MainTextSectionRef}[2]{\ifdef{\UseHardcodedReferences}{{Section~#1}}{Section~\ref{#2}}}
\newcommand{\MainTextEquationRef}[2]{\ifdef{\UseHardcodedReferences}{{Eq.~(#1) in the text}}{Eq.~(\ref{#2})}}
\newcommand{\MainTextFigureRef}[2]{\ifdef{\UseHardcodedReferences}{{Fig.~#1}}{Fig.~\ref{#2}}}
\newcommand{\MainTextTableRef}[2]{\ifdef{\UseHardcodedReferences}{{Table~#1 of the text}}{Table~\ref{#2}}}

\newcommand{\IfAnonymous}[2]{\ifdef{\UseAnonymousText}{#1}{#2}}

\newcommand{\CaptionToTableVspace}{
}
\newcommand{\AfterTableVspace}{\vspace{0.30in}}

\renewcommand{\changed}[1]{#1} 
\renewcommand{\moved}[1]{#1} %

\usepackage{mathtools}
\usepackage[capitalise]{cleveref}

\makeatletter
\DeclareRobustCommand\onedot{\futurelet\@let@token\@onedot}
\def\@onedot{\ifx\@let@token.\else.\null\fi\xspace}
\def\eg{\emph{e.g}\onedot} 
\def\ie{\emph{i.e}\onedot}

\let\@authorsaddresses\@empty
\makeatother

\let\savedbaselinestretch\baselinestretch
\begin{document}

\title
[\OurMethod{}: Streamable Memory Efficient Radiance Fields for Real-Time Large-Scene Exploration]
{\OurMethod{}: Streamable Memory Efficient Radiance Fields \\
 for Real-Time Large-Scene Exploration}

\IfAnonymous{}{
    \author{Daniel Duckworth}
    \authornote{Denotes equal contribution.}
    \orcid{0009-0002-8716-5934}
    \affiliation{%
      \institution{Google DeepMind}
      \country{Germany}
    }
    \email{duckworthd@gmail.com}
    
    \author{Peter Hedman}
    \authornotemark[1]
    \orcid{0000-0002-2182-0185}
    \affiliation{%
      \institution{Google Research}
      \country{United Kingdom}
    }
    \email{hedman@gmail.com}
    
    \author{Christian Reiser}
    \orcid{0009-0002-1050-3958}
    \affiliation{%
      \institution{Tübingen AI Center, University of Tübingen}
      \country{Germany}
    }
    \affiliation{%
      \institution{Google Research}
      \country{USA}
    }
    \email{christian.j.reiser@gmail.com}
    
    \author{Peter Zhizhin}
    \orcid{0009-0004-8850-252X}
    \affiliation{%
      \institution{Google Research}
      \country{Germany}
    }
    \email{pzhizhin@google.com}
    
    \author{Jean-François Thibert}
    \orcid{0009-0002-8095-4121}
    \affiliation{%
      \institution{Google Inc.}
      \country{Canada}
    }
    \email{jfthibert@google.com}
    
    \author{Mario Lučić}
    \orcid{0009-0000-7826-2340}
    \affiliation{%
      \institution{Google DeepMind}
      \country{Switzerland}
    }
    \email{lucic@google.com}
    
    \author{Richard Szeliski}
    \orcid{0009-0005-5300-5475}
    \affiliation{%
      \institution{Google Research}
      \country{USA}
    }
    \email{szeliski@google.com}
    
    \author{Jonathan T. Barron}
    \orcid{0009-0008-4016-9448}
    \affiliation{%
      \institution{Google Research}
      \country{USA}
    }
    \email{barron@google.com}
}

\begin{abstract}

Recent techniques for real-time view synthesis have rapidly advanced in fidelity and speed, and modern methods are capable of rendering near-photorealistic scenes at interactive frame rates.
At the same time, a tension has arisen between explicit scene representations amenable to rasterization and neural fields built on ray marching, with state-of-the-art instances of the latter surpassing the former in quality while being prohibitively expensive for real-time applications.
We introduce \OurMethod{}, a view synthesis approach that achieves state-of-the-art accuracy among real-time methods on large scenes with footprints up to 300 m\textsuperscript{2} at a volumetric resolution of 3.5 mm\textsuperscript{3}.
Our method is built upon two primary contributions: a hierarchical model partitioning scheme, which increases model capacity while constraining compute and memory consumption, and a distillation training strategy that simultaneously yields high fidelity and internal consistency.
Our method enables full six degrees of freedom navigation in a web browser and renders in real-time on commodity smartphones and laptops.
Extensive experiments show that our method exceeds the state-of-the-art in real-time novel view synthesis by 0.78 dB on standard benchmarks and 1.78 dB on large scenes, renders frames three orders of magnitude faster than state-of-the-art radiance field models, and achieves real-time performance across a wide variety of commodity devices, including smartphones.
\IfAnonymous{
}{
    We encourage readers to explore these models interactively at our project website: \OurWebsite{}.
}

\end{abstract}



\begin{teaserfigure}
    \centering
    \captionsetup{type=figure}
        \centering
    \IfAnonymous{
        \includegraphics[width=0.95\linewidth]{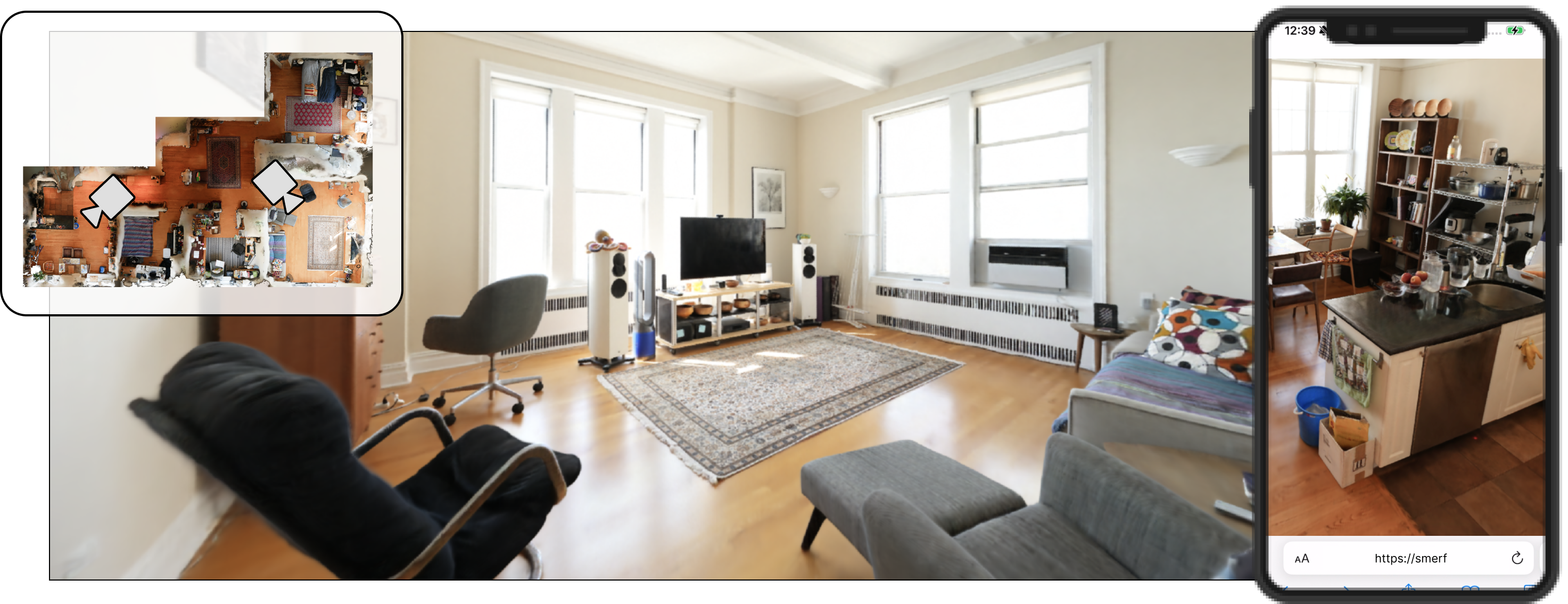}
    }{
        \includegraphics[width=0.95\linewidth]{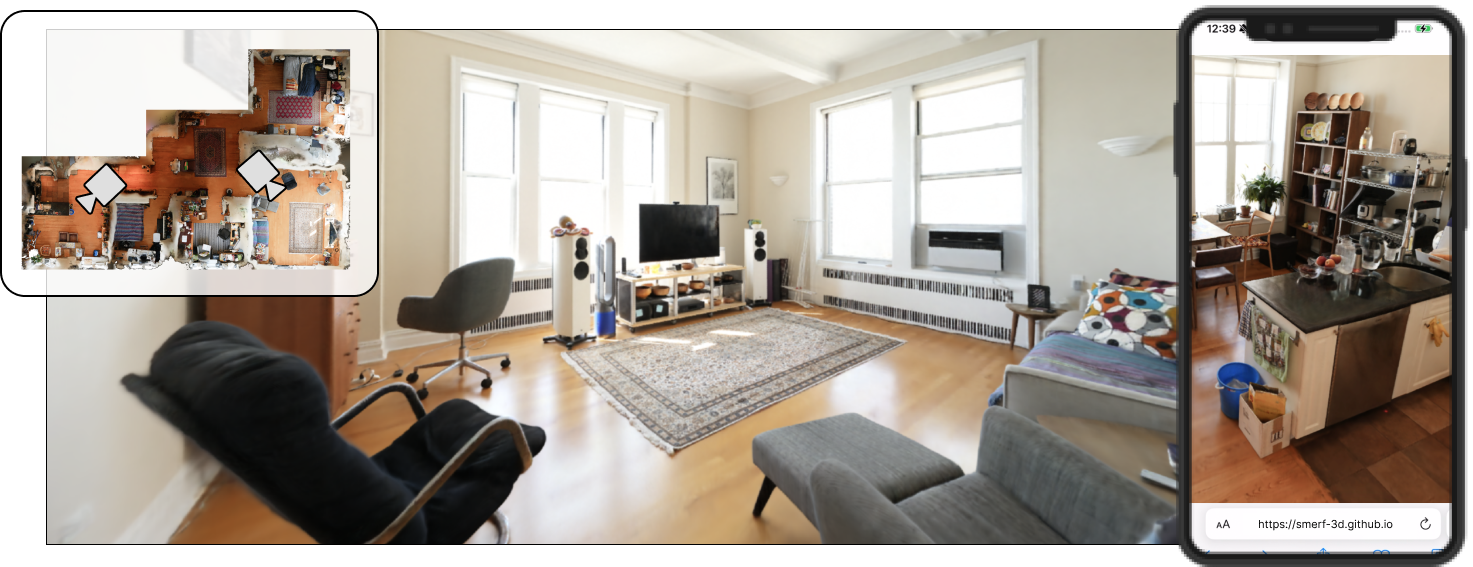}
    }
    \caption{
        \OurMethod{} achieves real-time view-synthesis of large scenes on commodity devices while approaching the quality of state-of-the-art offline methods.
        By streaming content based on viewpoint, our method scales to hundreds of m$^2$ and runs in the browser on resource-constrained devices, including smartphones.
    }
    \label{fig:teaser}

\end{teaserfigure}

\maketitle
    
\section{Introduction}
\label{sec:introduction}

Radiance fields have emerged as a powerful, easily-optimized representation for reconstructing and re-rendering photorealistic real-world 3D scenes.
In contrast to explicit representations such as meshes and point clouds, radiance fields are often stored as neural networks and rendered using volumetric ray-marching.
This is simultaneously the representation's greatest strength and its biggest weakness: neural networks can concisely represent complex geometry and view-dependent effects given a sufficiently large computational budget.
As a volumetric representation, the number of operations required to render an image scales in the number of pixels rather than the number of primitives (e.g. triangles), with the best-performing models~\cite{barron2023zipnerf} requiring tens of millions of network evaluations.
As a result, real-time radiance fields make concessions in quality, speed, or representation size, and it is an open question if they can compete with alternative approaches such as Gaussian Splatting~\cite{kerbl20233d}.

Our work answers this question in the affirmative.
We present a scalable approach to real-time rendering of large spaces at higher fidelity than previously possible.
Not only does our method approach the quality of slower, state-of-the-art models in standard benchmarks, it is the first to convincingly render unbounded, multi-room spaces in real-time on commodity hardware.
Crucially, our method achieves this with a memory budget independent of scene size without compromising either image quality or rendering speed.

We use a variant of Memory-Efficient Radiance Fields (\MERF{}) \cite{reiser2023merf}, a compact representation for real-time view synthesis, as a core building block.
We construct a hierarchical model architecture composed of self-contained \MERF{} submodels, each specialized for a region of viewpoints in the scene.
This greatly increases model capacity with bounded resource consumption, as only a \emph{single} submodel is necessary to render any given camera.
Specifically, our architecture tiles the scene with submodels to increase spatial resolution and also tiles parameters within each submodel region to more accurately model view-dependent effects.
We find the increase in model capacity to be a double-edged sword, as our architecture lacks the inductive biases of state-of-the-art offline models that encourage plausible reconstructions.
We therefore introduce an effective new distillation training procedure that provides abundant supervision of both color and geometry, including generalization to novel viewpoints and stable results under camera motion.
These two innovations enables the real-time rendering of large scenes with similar scale and quality to existing state-of-the-art work.

Concretely, our two main contributions are:

\begin{itemize}
    \item a tiled model architecture for real-time radiance fields capable of representing large scenes at high fidelity on hardware ranging from smartphones to desktop workstations.
    \item a radiance field distillation training procedure that produces highly consistent submodels with the generalization capabilities and inductive biases of an accurate but compute-heavy teacher.
\end{itemize}

\section{Related Work}
\label{sec:related_work}

Neural Radiance Fields have seen explosive progress, making a complete review challenging.
Here, we only review papers directly related to our work, either as building blocks or sensible alternatives.
Please see \citet{tewari2022advances} for a comprehensive overview.

\subsection{Improving NeRF's Speed}
Although NeRF produces compelling results, it is severely lacking in terms of speed: training a model takes multiple hours and rendering an image requires up to 30 seconds~\cite{mildenhall2020nerf}.
Since then, much effort has been placed on accelerating NeRF's training and inference.

Many works achieve real-time rendering by precomputing (\ie, \emph{baking}) NeRF's view-dependent colors and opacities into sparse volumetric data structures~\cite{hedman2021snerg,yu2021plenoctrees,garbin2021fastnerf,hyan2023plenvdb}.
Other works show that rendering and/or training can be significantly accelerated by parameterizing a radiance field with a dense~\cite{yu2021plenoxels,sun2022direct,wu2022diver}, hashed~\cite{muller2022instant} or low-rank voxel grid parameterization~\cite{chan2021eg3d,chen2022tensorf}, a grid of small MLPs~\cite{reiser2021kilonerf}, or a ``polygon soup'' of neural textures~\cite{chen2022mobilenerf}.
Alternatively, rendering can be accelerated by reducing the sample budget while carefully allocating volumetric samples to preserve quality~\cite{neff2021donerf,kurz2022adanerf,adaptiveshells2023,mcnerf2023}

In general, there is a tension between render time and memory consumption. This was explored in \MERF{}~\cite{reiser2023merf}, which combines sparse and low-rank voxel grid representations to enable real-time rendering of unbounded scenes within a limited memory budget~\cite{reiser2023merf}.
We build upon \MERF{} and extend it to much larger environments while retaining real-time performance and adhering to the tight memory constraints of commodity devices.

\subsection{Improving NeRF's Quality}
In parallel, the community has improved \NeRF{}'s quality in a variety of ways,
\eg by eliminating aliasing~\cite{barron2021mipnerf,hu2023tri},
introducing efficient model architectures~\cite{muller2022instant},
modeling unbounded scenes~\cite{barron2022mipnerf360},
and eliminating floaters~\cite{philip2023gradient}.
Other works improve robustness to other challenges,
such as inaccurate camera poses~\cite{lin2021barf,song2023sc,park2023camp},
limited supervision~\cite{Niemeyer2021Regnerf,roessle2023ganerf,wu2023reconfusion,yang2023freenerf},
or outliers including illumination changes,
transient objects~\cite{martinbrualla2020nerfw,rematas2022urban},
and motion~\cite{park2021nerfies,jiang2023alignerf}.

Most relevant for our work is \ZipNeRF{}~\cite{barron2023zipnerf}, which uses multisampling to enable anti-aliasing for fast grid-based representations
and produces high-quality reconstructions of large environments while remaining tractable to train and render.
Although \ZipNeRF{} currently achieves state-of-the-art quality on established benchmarks, a single high-resolution frame requires multiple seconds to render, leaving the method unsuitable for real-time rendering.
We therefore adopt the approach of distilling a high-fidelity \ZipNeRF{} model into a set of \MERF{}-based submodels, thereby achieving quality comparable to \ZipNeRF{} at real-time frame rates.
We also incorporate several quality improvement listed above, such as latent codes for illumination variation~\cite{martinbrualla2020nerfw} and gradient scaling for floaters removal~\cite{philip2023gradient}, with no impact to rendering speed.
More details can be found in \appendixref{E}{sec:teacher_training}.

\subsection{Rasterization-based View-Synthesis}
While \NeRF{} synthesises views using per-pixel ray-marching, an alternative paradigm is per-primitive rasterization leveraging specialized GPU hardware.
Early methods approximated geometry with triangle meshes and used image blending to model view-dependent appearance~\cite{debevec1998efficient,buehler01ulr,davis2012}.
Later methods improved quality with neural appearance models~\cite{hedman2018deep,martin2018lookingood,liu2023real} or neural mesh reconstruction~\cite{wang2021neus,bakedsdf,rojas2023re,rakotosaona2023nerfmeshing,philip2021free}.
Recent methods model detailed geometry by rasterizing overlapping meshes and either decoding~\cite{Wan_2023_CVPR} or compositing~\cite{chen2022mobilenerf} the results.
For limited viewing volumes, semi-transparency can be modeled with layered representations such as multi-plane images~\cite{Soft3DReconstruction,zhou2018stereomag,flynn2019deepview,mildenhall2019llff}.

GPU hardware also accelerates rendering of point-based representations.
This can be leveraged to synthesize views by decoding sparse point clouds with a U-Net~\cite{aliev2020neural,wiles2020synsin,ruckert2021adop}. While the output views tend to be inconsistent under camera motion, this can be addressed by extending each point into a disc~\cite{szeliski1992surface,pfister2000surfels} equipped with a soft reconstruction kernel~\cite{zwicker2001surface}.
Recent work showed that these soft point-based representations are amenable to gradient-based optimization and can effectively model semi-transparency as well as view-dependent appearance~\cite{kopanas2021point,zhang2022differentiable}.

Most relevant to our work is 3D Gaussian Splatting~\cite{kerbl20233d} (3DGS), which improves visual fidelity, simplifies initialization, reduces the run-time of soft point-based representations, and represents the current state-of-the-art in real-time view-synthesis.
While 3DGS produces high-quality reconstructions for many scenes, optimization remains challenging, and carefully chosen heuristics are required for point allocation.
This is particularly evident in large scenes, where many regions lack sufficient point density.
To achieve real-time training and rendering, 3DGS further relies on platform-specific features and low-level APIs.
While recent viewers can display 3DGS models on commodity hardware~\cite{splat_antimatter,gsplat_tech,gaussiansplats3d}, they rely on approximations to sort order and view-dependency whose impact on quality has not yet been evaluated. 
Concurrent work~\cite{fan2023lightgaussian,lee2023compact} suffers a small reduction in visual fidelity to significantly decrease model size.
In our experiments, we directly compare our cross-platform web viewer with the latest version of the official 3DGS viewer.

\subsection{Large Scale NeRFs}
Although NeRF models excel at reproducing objects and localized regions of a scene, they struggle to scale to larger scenes.
For object-centric captures, this can be ameliorated by reparameterizing the unbounded background regions of the scene to a bounded domain~\cite{kaizhang2020,barron2022mipnerf360}.
Larger, multi-room scenes can be modeled by applying anti-aliasing techniques to hash grid-backed radiance fields~\cite{barron2023zipnerf,VRNeRF}.
Another approach is to split the scenes into multiple regions and to train a separate NeRF for each region~\cite{rebain2021derf,turki2022meganerf,wu2023scanerf}.
This facilitates real-time rendering of both objects~\cite{reiser2021kilonerf} and room-scale scenes~\cite{wu2022snisr}.
Scaling to extremely large scenes such as city blocks or entire urban environments, however, requires partitioning based on camera location into redundant, overlapping scene volumes~\cite{tancik2022blocknerf,meuleman2023localrf}.
We extend the idea of camera-based partitioning to real-time view-synthesis: while existing models require expensive rendering and image blending from multiple submodels, we need only one submodel to render a given camera and leverage regularization during training to encourage mutual consistency.

\subsection{Distillation and NeRF}
A powerful concept in deep learning is that of \emph{distillation} --- training a smaller or more efficient \emph{student} model to approximate the output of a more expensive or cumbersome \emph{teacher}~\cite{hinton2015distilling,gou2021knowledge}.
This idea has been successfully applied to NeRFs in a variety of contexts such as (i) distilling a large MLP into a grid of tiny MLPs~\cite{reiser2021kilonerf}, (ii) distilling an expensive NeRF MLP into a small ``proposal'' MLP that bounds density~\cite{barron2022mipnerf360}, or (iii) distilling expensive secondary ray bounces into lightweight model for inverse rendering~\cite{nerv2021}.
Distillation has also been used to avoid expensive ray marching and facilitate real-time rendering by converting an entire NeRF scene into a light field model~\cite{wang2022r2l,attal2022learning,cao2023real,gupta2023lightspeed} or a surface representation~\cite{rakotosaona2023nerfmeshing}.

In this work, we distill the appearance and geometry of a large, high-quality \ZipNeRF{} model into a family of \MERF{}-like submodels.
Concurrently, HybridNeRF~\cite{turki2023hybridnerf} also employs distillation for real-time view synthesis, albeit for a signed distance field.
    
\section{Preliminaries}
\label{sec:preliminaries}

We begin with a review of MERF, which maps 3D positions $\PPosition\in\mathbb{R}^3$ to feature vectors $\mathbf{t}\in\mathbb{R}^8$. MERF parameterizes this mapping using a combination of high-resolution triplanes ($\TriplaneFeatureYZ$, $\TriplaneFeatureXZ$, $\TriplaneFeatureXY\in\Real^{\TriplaneResolution\TimesNoWS\TriplaneResolution\TimesNoWS 8}$) and a low-resolution sparse voxel grid $\SparseGridFeature\in\Real
^{\SparseGridResolution\TimesNoWS\SparseGridResolution\TimesNoWS\SparseGridResolution\TimesNoWS 8}$. A query point $\PPosition$ is projected onto each of three axis-aligned planes and the underlying 2D grid is queried via bilinear interpolation. Additionally, a trilinear sample is taken from the sparse voxel grid. The resulting four 8-dimensional vectors are then summed:
\begin{equation}\label{eq:merf_aggregation}
    \PositionFeature_{\text{MERF}}(\PPosition) = \TriplaneFeatureYZ(\PPosition)+\TriplaneFeatureXZ(\PPosition)+\TriplaneFeatureXY(\PPosition)+\SparseGridFeature(\PPosition) \,.
\end{equation}
This vector is then unpacked into three parts, which are independently rectified to yield a scalar density, a diffuse RGB color, and a feature vector that encodes view-dependence effects:
\begin{equation}
    \density = \exp(\mathbf{t}_1)\,,\quad 
    \mathbf{c} = \operatorname{sigmoid}(\mathbf{t}_{2:4})\,,\quad
    \mathbf{f} = \operatorname{sigmoid}(\mathbf{t}_{5:8})\,.
\end{equation}
To render a pixel, a ray is cast from that pixel's center of projection $\mathbf{o}$ along the viewing direction $\mathbf{d}$ and sampled at a set of distances $\{ t_i \}$ to generate a set of points along that ray $\mathbf{x}_i = \mathbf{o} + t_i\mathbf{d}$.
As in NeRF~\cite{mildenhall2020nerf}, the densities $\{\density_i\}$ are converted into alpha compositing weights $\{ w_i \}$ using the numerical quadrature approximation for volume rendering~\cite{max1995optical}:
\begin{equation} \label{eq:nerfcolor}
    w_i = T_i (1 - \exp(-\density_i \delta_i)) \mbox{, where }
    T_i = \exp \lft( -\sum\limits_{j < i} \density_j \delta_j \rgt) \, ,
\end{equation}
where $\delta_i = t_{i+1} - t_i$ is the distance between adjacent samples.
After alpha compositing, the deferred shading approach of  \SNeRG{}~\cite{hedman2021snerg} is used to decode the blended diffuse RGB colors $\sum_{i} w_i \mathbf{c}_{i}$ and the blended view-dependent color features $\sum_{i} w_i \mathbf{f}_{i}$ into the final pixel color with the help of the small deferred rendering MLP $h(\cdot; \DeferredMlpParams)$:
\begin{equation} \label{eq:merf_deferred}
\lft( \sum_{i} w_i \mathbf{c}_{i} \rgt) + h\lft(\sum_{i} w_i \mathbf{c}_{i},\,\sum_{i} w_i \mathbf{f}_{i},\,\mathbf{d}; \DeferredMlpParams \rgt)\,,
\end{equation}
where $\DeferredMlpParams$ are the MLP's parameters.

In unbounded scenes, far-away content can be modelled coarsely. To achieve a resolution that drops off with the distance from the scene's focus point, MERF applies a contraction function to each spatial position $\mathbf{x}$ before querying the feature field:
\begin{equation} 
\operatorname{contract}(\mathbf{x})_d = \begin{cases}
x_d &\text{if } \|\mathbf{x}\|_\infty \leq 1\\
\frac{x_d}{\prescript{}{\phantom{d}}{\|\mathbf{x}\|_\infty}} &\text{if } x_d \neq \|\mathbf{x}\|_\infty > 1 \\
\left(2 - \frac{1}{|x_d|}\right) \frac{x_d}{|x_d|} & \text{if } x_d = \|\mathbf{x}\|_\infty > 1\end{cases}
\label{eqn:contract360}
\end{equation}

\section{Model}
\label{sec:method}
\begin{figure*}[t]
    \newcommand{\CameraColor}[1]{{\color[HTML]{CC1A14}{\textbf{#1}}}}
    \newcommand{\ScenePartitionColor}[1]{{\color[HTML]{3C8FBB}{\textbf{#1}}}}
    \newcommand{\SubmodelOneColor}[1]{{\color[HTML]{2A5274}{\textbf{#1}}}}
    \newcommand{\SubmodelTwoColor}[1]{{\color[HTML]{FF7F0E}{\textbf{#1}}}}
    \newcommand{\TrilerpColor}[1]{{\color[HTML]{3DA900}{\textbf{#1}}}}

    \centering
    \includegraphics[trim={0cm 28.8cm 40.5cm 0cm},clip,width=0.82\linewidth]{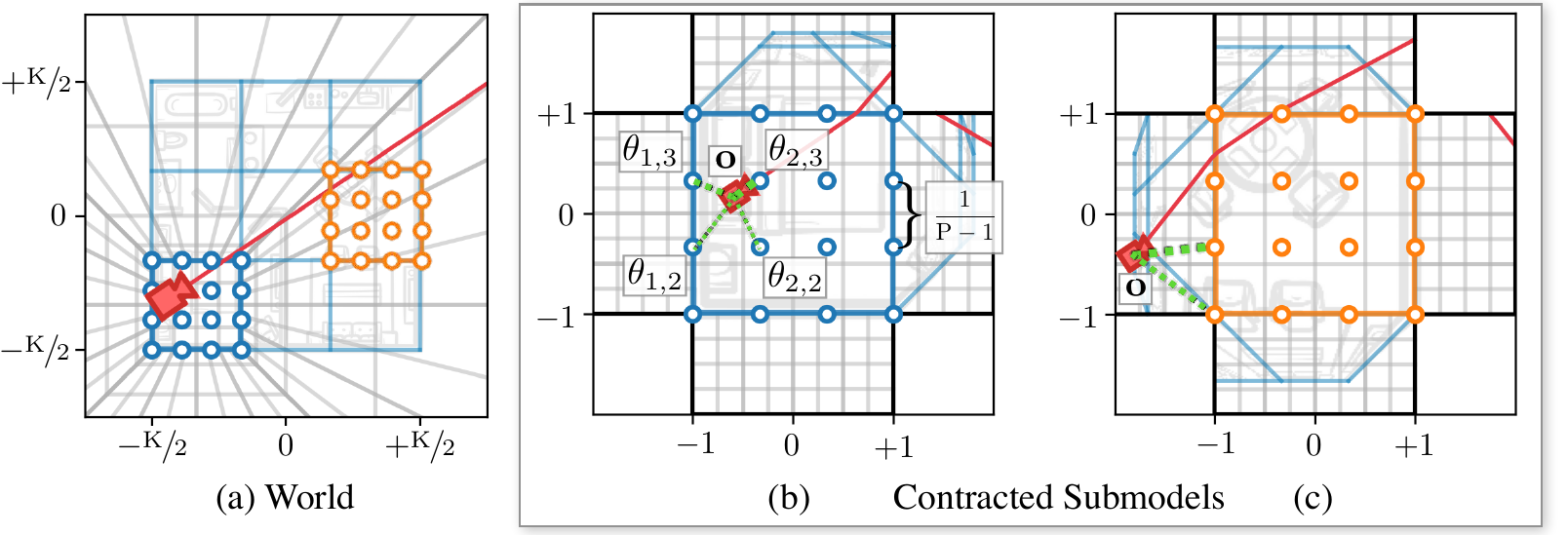}
    \caption{
        \textbf{Coordinate systems} in \OurMethod{} for a scene with $\SubmodelResolution^3 = 3^3$ coordinate space partitions and $\DeferredResolution^3 = 4^3$ deferred appearance network sub-partitions.
        Each partition is capable of representing the entire scene while allocating the majority of its model capacity to its corresponding partition.
        Within each partition, we instantiate a set of spatially-anchored MLP weights $\{ \theta_{i,j} \}$ parameterizing the deferred appearance model, which we \TrilerpColor{trilinearly interpolate} as a function of the \CameraColor{camera} origin $\CameraOrigin$ during rendering.
        In (a), we present the entire scene in world coordinates with the \ScenePartitionColor{scene partition} and highlight \SubmodelTwoColor{two} \SubmodelOneColor{submodels}.
        In (b) and (c) we present the same scene from the view of two submodels in their corresponding contracted coordinate systems.
        (b) visualizes the rendering and parameter interpolation process when the camera origin $\CameraOrigin$ lies \SubmodelOneColor{inside} of a submodel's partition, and (c) visualizes the same when it lies \SubmodelTwoColor{outside}.
    }
    \label{fig:coordinate_systems}
\end{figure*}
Although real-time view-synthesis methods like \MERF{} perform well for a localized environment, they often fail to scale to large multi-room scenes.
To this end, we present a hierarchical architecture.
First, we partition the coordinate space of the entire scene into a series of blocks, where each block is modeled by its own \MERF{}-like representation.
Second, we introduce a grid of spatially-anchored network parameters \emph{within} each block for modeling view-dependent effects.
Finally, we introduce a gating mechanism for modulating high- and low-resolution contributions to a location's feature representation.
Our overall architecture can be thought of as a three-level hierarchy: based on the camera origin,
(i) we select an appropriate submodel, then within a submodel 
(ii) we compute the parameters of a deferred appearance network via interpolation, and then within a local voxel neighborhood,
(iii) we determine a location's feature representation via feature gating.

This greatly increases the capacity of our model without diminishing rendering speed or increasing memory consumption: even as total storage requirements increase with the number of submodels, only a \emph{single} submodel is needed to render a given frame.
As such, when implemented on a graphics accelerator, our system maintains modest resource requirements comparable to \MERF{}.

\subsection{Coordinate Space Partitioning}
\label{sec:coordinate-space-partitioning}
While \MERF{} offers sufficient capacity for faithfully representing medium-scale scenes, we found the use of a single set of triplanes limits its capacity and reduces image quality.
In large scenes, numerous surface points project to the same 2D plane location, and the representation therefore struggles to simultaneously represent high-frequency details of multiple surfaces.
Although this can be partially ameliorated by increasing the spatial resolution of the underlying representation, doing so significantly impacts memory consumption and is prohibitively expensive in practice.

We instead opt to coarsely subdivide the scene into a 3D grid based on camera origin and associate each grid cell with an independent submodel in a strategy akin to \BlockNeRF~\cite{tancik2022blocknerf}.
Each submodel is assigned its own contraction space (\cref{eqn:contract360}) and is tasked with representing the region of the scene within its grid cell at high detail, while the region outside each submodel's cell is modeled coarsely.
Note that the entire scene is still represented by each submodel --- the submodels differ only in terms of which portions of the scene lies inside or outside of each submodel's contraction region.
As such, rendering a camera only requires \emph{a single submodel}, implying that only one submodel must be in memory at a time.

Formally, we shift and scale all training cameras to lie within a $[-\sfrac{\SubmodelResolution}{2}, \sfrac{\SubmodelResolution}{2}]^3$ cube, and then partition this cube into $\SubmodelResolution^3$ identical and tightly packed subvolumes of size $[-1, 1]^3$.
We assign training cameras to submodels $\{ \Submodel{}_{\SubmodelIndex{}} \}$ by identifying the associated subvolume $\FocusRegion_{\SubmodelIndex{}}$ that each camera origin $\CameraOrigin$ lies within:
\begin{equation}
    \label{eqn:submodel-assign}
    s^{*} 
        = \argmin_{\SubmodelIndex{} \in \{ 1..\SubmodelResolution{}^3 \} } \DistanceToSubmodel(\CameraOrigin, \SubmodelIndex{})
    \mathrm{, where }\:
    \DistanceToSubmodel(\CameraOrigin, \SubmodelIndex{})
        = \min_{\PPosition \in \FocusRegion_{\SubmodelIndex{}}} || \CameraOrigin - \PPosition ||_{\infty}.
\end{equation}
We design our camera-to-submodel assignment procedure to apply to cameras outside of the training set, ensuring its validity during test set rendering.
This enables a wide range of features including ray jittering, submodel reassignment, a submodel consensus loss, arbitrary test camera placement, and ping-pong buffers as described in Sections~\ref{sec:training} and \ref{sec:rendering}.

Rather than exhaustively instantiating submodels for all $\SubmodelResolution^3$ subvolumes, we only consider subvolumes that contain at least one training camera.
As most scenes are outdoors or single-story buildings, this reduces the number of submodels from $\SubmodelResolution{}^3$ to $O(\SubmodelResolution{}^2)$.

\subsection{Deferred Appearance Network Partitioning}
\label{sec:deferred-appearance-network-partitioning}
The second level in our partitioning hierarchy concerns the deferred rendering model.
Recall that \MERF{} employs a small multi-layer perceptron (MLP) to decode view-dependent colors from blended features as described in \cref{eq:merf_deferred}.
Although the small size of this network is critical for fast inference, we observed that its capacity is insufficient to accurately reproduce complex view-dependent effects common in larger scenes.
Simply increasing the size of this network is not viable as doing so would significantly reduce rendering speed.

Instead, we uniformly subdivide the domain of each submodel into a lattice with $\DeferredResolution$ vertices along each axis.
We associate each cell $(u,v,w)\in\{1,..,\DeferredResolution\}^3$ with a separate set of network parameters $\DeferredMlpParams_{uvw}$ and trilinearly interpolate them based on camera origin $\CameraOrigin$:
\begin{equation}
    \DeferredMlpParams = \Trilerp(
        \CameraOrigin, 
        \{ \DeferredMlpParams_{uvw}: (u,v,w)\in\{1,..,\DeferredResolution\}^3\}
    )
\end{equation}
The use of trilinear interpolation, unlike the nearest-neighbor interpolation used in coordinate space partitioning, is critical in preventing aliasing of the view-dependent MLP parameters, which takes the form of conspicuous ``popping'' artifacts in specular highlights as the camera moves through space.
We further reduce popping \emph{between} submodels via regularization in Sec. \ref{sec:submodel_consistency}.
After parameter interpolation, view-dependent colors are decoded according to \cref{eq:merf_deferred}.

Since the size of the view-dependent MLP is negligible compared to total representation size, deferred network partitioning has almost no effect on memory consumption or storage impact.
As a result, this technique increases model capacity almost for free.
This is in contrast to coordinate space partitioning, which significantly increases storage size.
We find that the union of coordinate space and deferred network partitioning is critical for effectively increasing spatial and view-dependent resolution, respectively.
See \cref{fig:coordinate_systems} for an illustration of these two forms of partitioning.

\subsection{Feature Gating}
\label{sec:feature-gating}
The final level in our hierarchy is at the level of \MERF{}'s coarse 3D voxel grid $\SparseGridFeature$ and three high-resolution planes ($\TriplaneFeatureYZ$, $\TriplaneFeatureXZ$, $\TriplaneFeatureXY$).
In \MERF{}, each 3D position is associated with an 8-dimensional feature vector: the sum of the contributions from these four sources (\cref{eq:merf_aggregation}).
Although effective, the features generated by this procedure are limited by their naive use of summation to merge high- and low-resolution information, entangling the two together.

We instead elect to use low-resolution 3D features to ``gate'' high-resolution features: if high-resolution features add value for a given 3D coordinate, they should be employed; otherwise, they should be ignored and the smoother, low-resolution features should be used.
To this end, we modify feature aggregation as follows: instead of a naive summation, we take the last component $\FeatureGatingWeight(\PPosition) = [\SparseGridFeature(\PPosition)]_{8}$ of the low-resolution voxel grid's contribution and use it to scale the triplane feature contributions:
\begin{equation}
    \hat{\PositionFeature}(\PPosition) =
        \FeatureGatingWeight(\PPosition)
        \cdot \left(
            \TriplaneFeatureYZ(\PPosition)
            + \TriplaneFeatureXZ(\PPosition)
            + \TriplaneFeatureXY(\PPosition)
        \right)
        + \SparseGridFeature(\PPosition)\,.
\end{equation}
We then build our final feature representation by concatenating the aggregated features $\hat{\PositionFeature}(\PPosition)$ and the voxel grid features $\SparseGridFeature(\PPosition)$:
\begin{equation}
    \PositionFeature(\PPosition) = \hat{\PositionFeature}(\PPosition) \oplus \SparseGridFeature(\PPosition)\,.
\end{equation}
Intuitively, this incentivizes the model to leverage the low-resolution voxel grid to disable the high-resolution triplanes when rendering low-frequency content such as featureless white walls, and gives the model the freedom to focus on detailed parts of the scene.
This change can also be thought of as a sort of ``attention'', as a multiplicative interaction is being used to determine when the model should ``attend'' to the triplane features~\cite{vaswani2017attention}.
This change slightly affects the memory and speed of our model by virtue of increasing the number of rows in the first weight matrix of our MLP, but the practical impact of this is negligible.
\section{Training}
\label{sec:training}

\subsection{Radiance Field Distillation}
\label{sec:distillation}
NeRF-like models such as MERF are traditionally trained ``from scratch'' to minimize photometric loss on a set of posed input images.
Regularization is critical when training such systems to improve generalization to novel views.
One example of this is \ZipNeRF{}~\cite{barron2023zipnerf}, which employs a family of carefully tuned losses in addition to photometric reconstruction error to achieve state-of-the-art performance on large, multi-room scenes.
We instead adopt ``student/teacher'' \emph{distillation} and train our representation to imitate an already-trained, state-of-the-art radiance field model.
In particular, we use a higher-quality variant of \ZipNeRF{}; see \appendixref{E}{sec:teacher_training}.

Distillation has several advantages: we inherit the helpful inductive biases of the teacher model,
circumvent the need for onerous hyperparameter tuning for generalization, and enable the recovery of local representations that are also globally consistent.
We show that our approach achieves quality comparable to its \ZipNeRF{} teacher while being three orders of magnitude faster to render.

\begin{figure}
    \centering
    \includegraphics[width=0.85\linewidth]{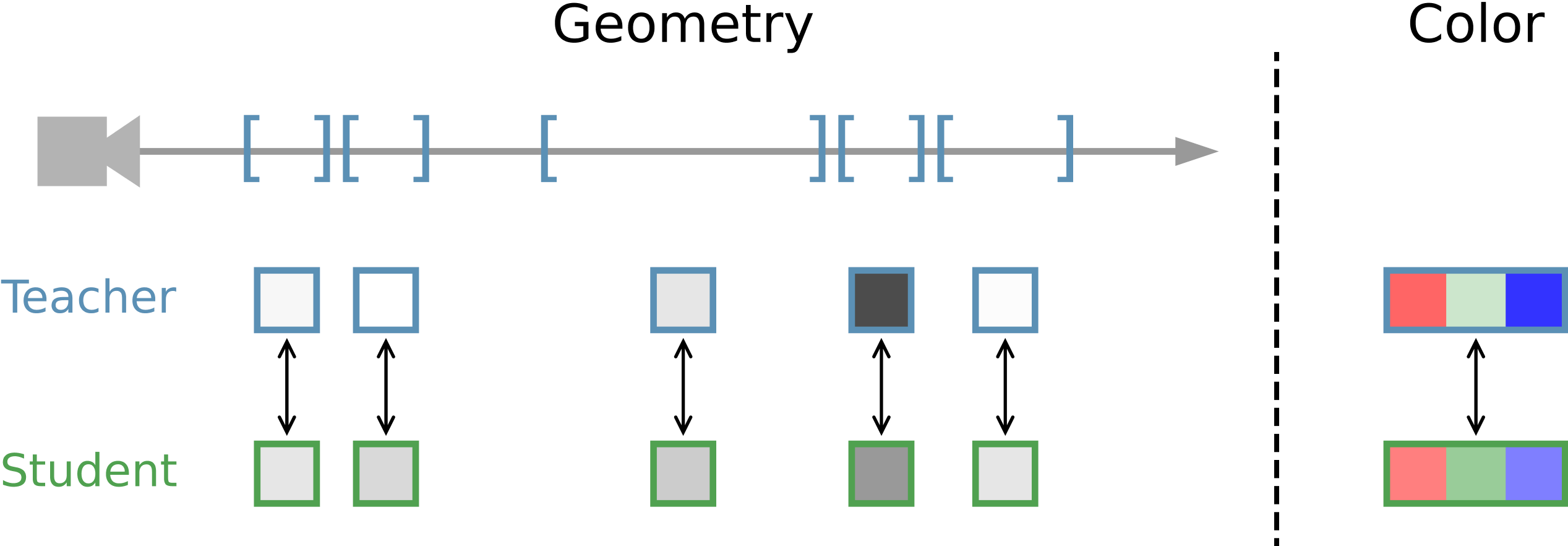}
    \caption{
        \textbf{Teacher Supervision}. 
        The student receives photometric supervision via rendered colors and geometric supervision via the rendering weights along camera rays.
        Both models operate on the same set of ray intervals.
    }
    \label{fig:geometry_supervision}
\end{figure}

We supervise our model by distilling the appearance and geometry of a reference radiance field.
Note that the teacher model is frozen during optimization. See \cref{fig:geometry_supervision} for a visualization.

\subsubsection{Appearance}
Like prior work, we supervise our model by minimizing the photometric difference between patches predicted by our model and a source of ground-truth.
Instead of limiting training to a set of photos representing a small subset of a scene's plenoptic function, our source of ``ground-truth'' image patches is a teacher model rendered from an \emph{arbitrary} set of cameras.
Specifically, we distill appearance by penalizing the discrepancy between $3\TimesNoWS3$ patches rendered from student and teacher models.
We use a weighted combination of the \rmse{} and \dssim{}~\cite{baker2023dssim} losses between each student patch $\mathbf{C}$ and its corresponding teacher patch $\mathbf{C}^*$:
\begin{align}
    \LossFn_{\radiance} =
        1.5 \cdot \dssim(\PatchRadiance, \PatchRadiance^*) + \sum_{\radiance \in \PatchRadiance} \norm{\radiance - \mathbf{\radiance}^*}_2 \,.
    \label{eqn:photometric-loss}
\end{align}
\subsubsection{Geometry}
To distill geometry, we begin by querying our teacher with a given ray origin and direction.
This yields a set of weighted intervals along the camera ray $\{((t_i, t_{i+1}), w^T_i)\}$, where each $(t_i, t_{i+1})$ are the metric distances along the ray corresponding to interval $i$, and each $w^T_i$ is the teacher's corresponding alpha compositing weight for the same interval as per \cref{eq:nerfcolor}.
The weight of each interval $i$ reflects its contribution to the final predicted radiance.
It is this quantity we distill into our student.
Specifically, we compute the absolute difference between the teacher and student weights: 
\begin{align}
    L_\density = \sum_{i} \left| w_i^T - w_i^S \right|\,.
    \label{eqn:geometry-reconstruction}
\end{align}
Because these volumetric rendering weights are a function of volumetric density (via \cref{eq:nerfcolor}), this loss on weights indirectly encourages the student's and teacher's density fields to be consistent with each other in visible regions of the scene.

\subsection{Data Augmentation}
\label{sec:regularization}
\begin{figure}[t]
    \centering
    \includegraphics[width=0.85\columnwidth]{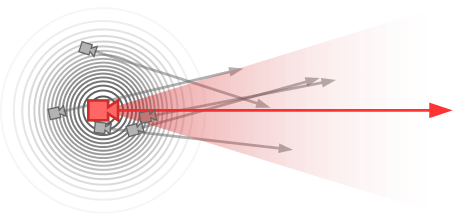}
    \caption{
        \textbf{Ray jittering}.
        To generate training rays for our student model (in gray) we randomly perturb the origins and directions of the camera rays used to supervise our teacher model (in red).
    }
    \label{fig:ray_jittering}
\end{figure}
Because our distillation approach enables the supervision of our student model at \emph{any} ray in Euclidean space, we require a procedure for selecting a useful set of rays.
While sampling rays uniformly at random throughout the scene is viable, this approach leads to poor reconstruction quality as many such rays originate from inside of objects or walls or are pointed towards unimportant or under-observed parts of the scene.
Using camera rays corresponding to pixels in the dataset used to train the teacher model also performs poorly, as those input images represent a tiny subset of possible views of the scene.
As such, we adopt a compromise approach by using randomly-perturbed versions of the dataset's camera rays, which yields a kind of ``data augmentation'' that improves generalization while focusing the student model's attention towards the parts of the scene that the photographer deemed relevant.

To generate a training ray, we first randomly select a ray from the teacher's dataset with origin $\origin$ and direction $\viewdir$.
We then jitter the origin with isotropic Gaussian noise and draw a uniform sample from an $\epsilon$-neighborhood of the ray's direction vector to obtain a ray $(\hat{\origin}, \hat{\viewdir})$ via the E3x library~\cite{unke2024e3x}: 
\begin{align}
    \hat{\origin} &\sim \Normal(\origin, \sigma^2 \Identity))\,, \\
    \hat{\viewdir} &\sim \Uniform(\{
        \mathbf{v} \in \Reals^3:
            \norm{\mathbf{v}-\viewdir}_2 < \epsilon,
            \norm{\mathbf{v}}_2=1
    \})\,.
\end{align}
In all experiments, we set $\sigma=0.03 \SubmodelResolution$ and $\epsilon=0.03$.
Note that $\sigma$ is defined in normalized scene coordinates, where all the input cameras are contained within the $[- \SubmodelResolution/2,  \SubmodelResolution/2]^3$ cube; See \cref{fig:ray_jittering}.

\subsection{Submodel Consistency}
\label{sec:submodel_consistency}

Recall that, in spite of employing a single teacher model, coordinate space partitioning means that we are effectively training multiple, independent student submodels in parallel (in practice, all submodels are trained simultaneously on a single host).
This presents a challenge in terms of \emph{consistency} across submodels --- at test time, we seek temporal consistency under smooth camera motion, even when transitioning between submodels.
This can be ameliorated by rendering multiple submodels and blending their results~\cite{tancik2022blocknerf}, but doing so significantly slows rendering and requires the presence of multiple submodels at once.
In contrast, we aim to render each frame while only querying a \emph{single} submodel.

To encourage adjacent submodels to make similar predictions for a given camera ray, we introduce a photometric consistency loss between submodels.
During training we render each camera ray in our batch twice: once using its ``home'' submodel $\submodel$ (whichever submodel the ray origin lies within the interior of), and again using a randomly-chosen neighboring submodel $\hat{\submodel}$.
We then impose a straightforward loss between those two rendered colors:
\begin{align}
    \LossFn_{\submodel} &=
        ||\radiance_{\submodel}(\Ray) - \radiance_{\hat{\submodel}}(\Ray)||_2\,.
    \label{eqn:submodel-consistency-loss}
\end{align}
Additionally, when constructing batches of training rays, we take care to assign each ray to a submodel where it will meaningfully improve reconstruction quality.
Intuitively, we expect rays to add the most value to their ``home'' submodel, but the rays that originate from neighboring submodels may also provide value by providing additional viewing angles of scene content within a submodel's interior.
As such, we first assign each training ray its ``home'' submodel, and then randomly re-assign $20\%$ of rays per batch to a randomly-selected adjacent submodel.

\section{Rendering}
\label{sec:rendering}
\subsection{Baking}
After training we generate precomputed assets for the real-time viewer. 
We broadly follow the ``baking'' process of \MERF{}~\cite{reiser2023merf} with the changes described here and other minor modifications in \appendixref{C}{sec:student_baking}.
\moved{Recall that \MERF{} extracts a multi-resolution occupancy grid for two purposes: (i) to mark the parts in the scene for baking, and (ii) to skip empty space during rendering.
We eliminate spurious floaters by post-processing this occupancy grid (at the highest resolution $\TriplaneResolution^3 = 2048^3$) with a 3x3x3 median filter.}
As output, we produce an \emph{independent} set of baked assets for each submodel, where each asset collection closely mirrors that produced by \MERF{}.
These assets include three high-resolution 2D feature maps and a sparse low-resolution 3D feature grid, both represented as quantized byte arrays.
The deferred network parameters, in contrast, retain their floating point representation.
Unlike \MERF{}, we store assets as gzip-compressed binary blobs, which we found slightly smaller and significantly faster to decode than PNG images.

\subsection{Live Viewer}
\label{sec:live-viewer}
\moved{
    Our viewer is based on the \MERF{} volume renderer~\cite{reiser2023merf}, \ie an OpenGL fragment shader that implements both ray marching and deferred rendering, using texture look-ups to retrieve pointwise feature representations and density. 
    However, our implementation contains several crucial modifications: support for submodels and parameter interpolation, a distance grid acceleration structure (\appendixref{D.1}{sec:distance-grid}), and other low-level optimizations (\appendixref{D.2}{sec:low-level-optimizations}).
}
We take particular care to ensure modest compute and memory requirements, enabling real-time rendering on smartphones and other resource-constrained devices.

\moved{
\subsubsection{Submodels}
Recall from \cref{sec:coordinate-space-partitioning} that our method only requires a \emph{single} submodel to render any given viewpoint in the scene, strongly bounding peak memory usage.
To hide network latency, we nevertheless ``ping-pong'' two submodels in and out of GPU memory as the user interactively explores the environment.
When the camera enters a new subvolume, we begin loading the corresponding submodel into CPU memory while the previous submodel, which includes an approximate representation of this new adjacent region, continues to be used for rendering.
Once the new submodel is available, we transfer it to GPU memory and immediately use it for rendering.
Peak GPU memory usage is thus limited to a maximum of two submodels: one actively being rendered and one being loaded into memory.
We further limit CPU memory usage by evicting loaded submodels from memory via least-recently-used caching.

\subsubsection{Deferred Appearance Network Interpolation}
While trilinearly interpolating deferred appearance network parameters requires additional computation, this feature has a negligible impact on frame rendering time.
As all pixels in an image share a common ray origin, we only need to interpolate parameters once per frame.
In practice, we perform this interpolation on the CPU before executing the fragment shader.
}

\begin{figure*}
    \centering
    \includegraphics[width=\linewidth]{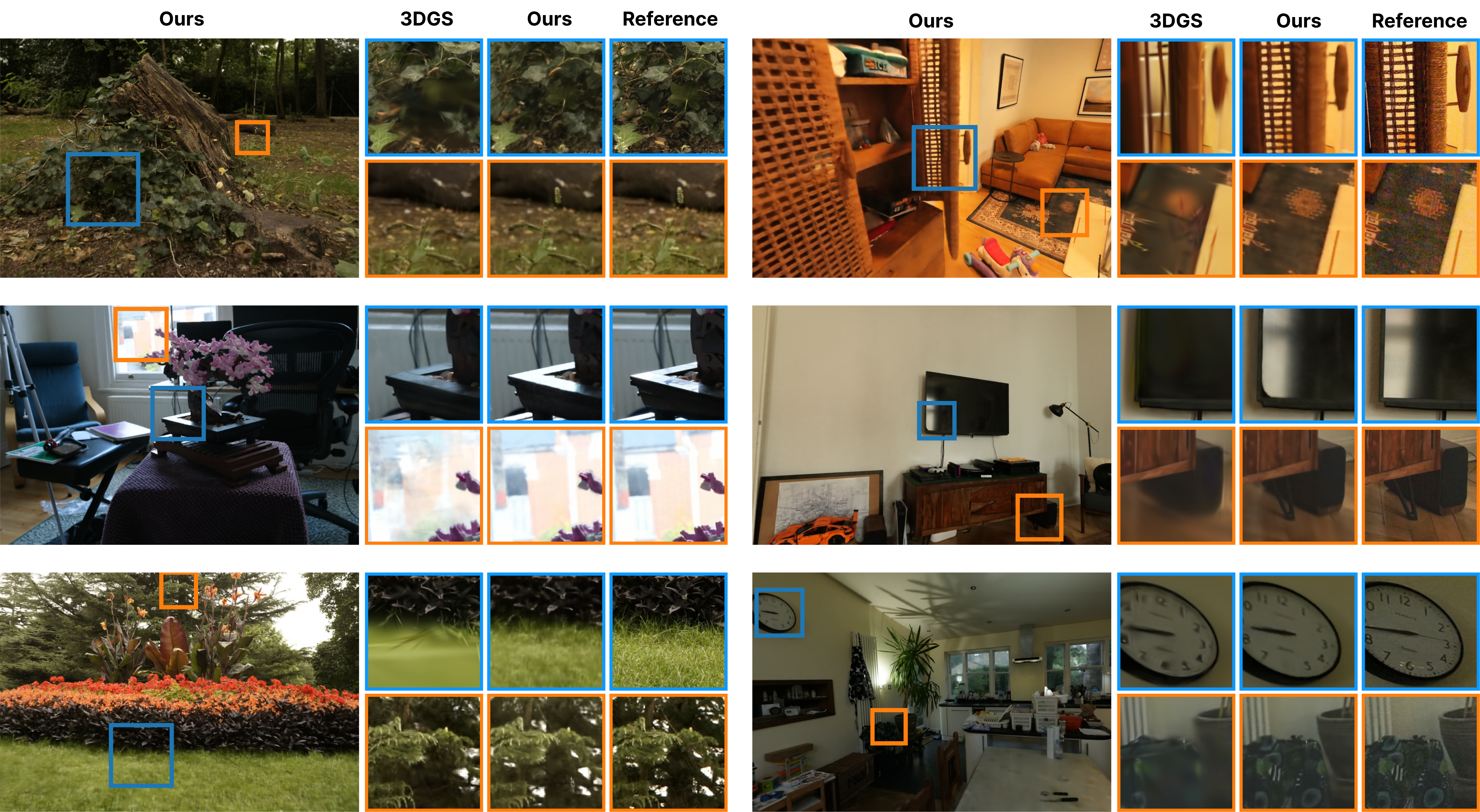}
    \caption{
        \textbf{Qualitative comparison}.
        We show results from our model and from 3D Gaussian Splatting~\cite{kerbl20233d} alongside ground-truth images on scenes from the mip-NeRF 360~\cite{barron2022mipnerf360} (left) and Zip-NeRF~\cite{barron2023zipnerf} (right) datasets.
        3D Gaussian Splatting struggles to reproduce the thin geometry, high-frequency textures, and view-dependent effects which our model successfully recovers.
    }
    \label{fig:vs_3dgs}
\end{figure*}
\begin{figure*}
    \centering
    \includegraphics[width=\linewidth]{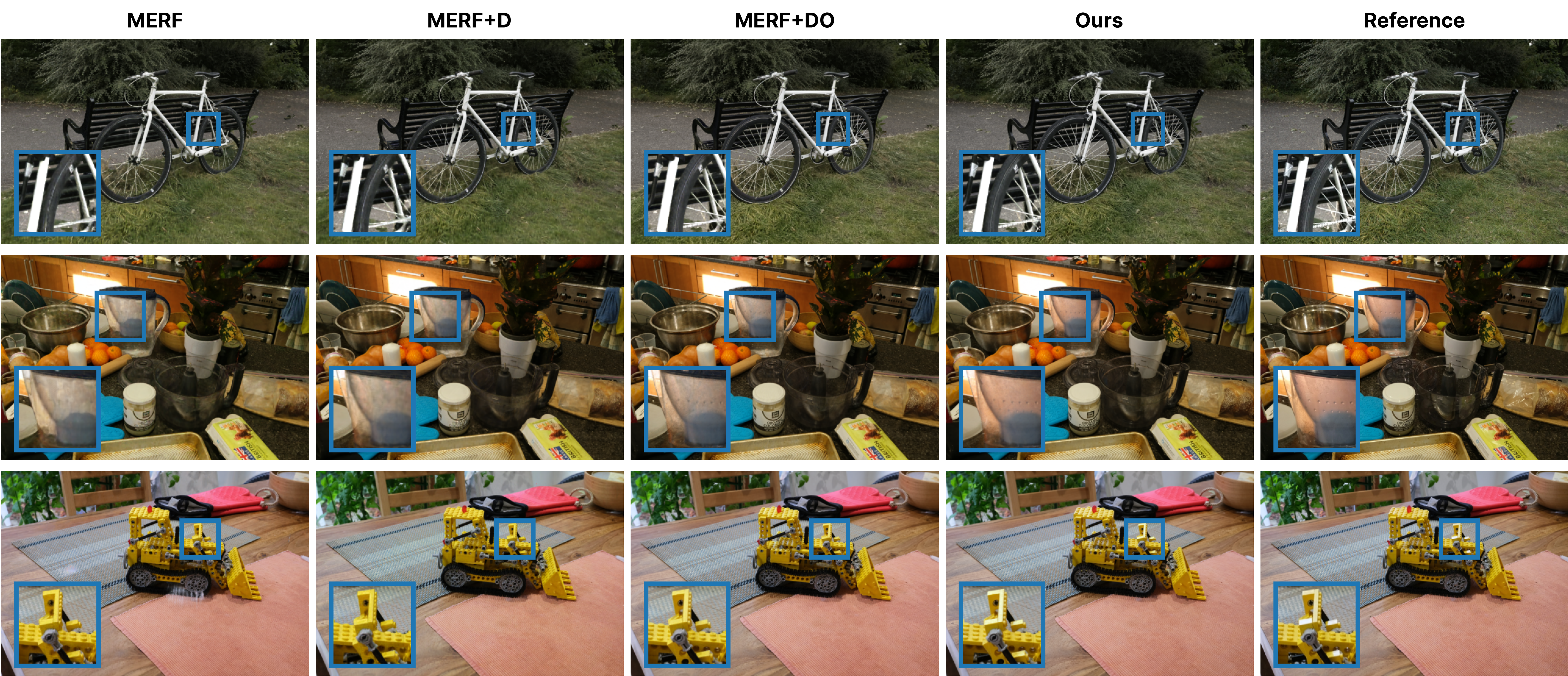}
    \caption{
        \textbf{Feature ablations}.
        We incrementally add distillation (\MERFDistill{}), optimization (\MERFDistillTrain{}), and model contributions (Ours) from Table~\ref{tab:mipnerf360_ablations} to \MERF{} to reach our submodel architecture.
        Distillation and optimization contributions markedly increase geometric and texture detail while model contributions improve view-dependent modeling accuracy.
    }
    \label{fig:vs_merf}
\end{figure*}

\begin{figure*}
    \input{figures/submodel_resolution}
\end{figure*}
\begin{figure}
    \input{figures/vs_zipnerf}
\end{figure}
\begin{figure}
    \input{figures/submodel_consistency}
\end{figure}
\begin{figure}
    \input{figures/ray_jittering_example}
\end{figure}

\section{Experiments}
\label{sec:experiments}

We evaluate our model's performance and quality by comparing it with two state-of-the-art methods: 
3D Gaussian Splatting~\cite{kerbl20233d} and \ZipNeRF{}~\cite{barron2023zipnerf}.
\ZipNeRF{} produces the highest-quality reconstructions of any radiance field model but is too slow to be useful in real-time contexts, while \GSplat{} renders quickly but reaches a lower quality than \ZipNeRF{}.
We show that our model,
(i) is comparable to \GSplat{} in runtime on powerful hardware, 
(ii) runs in real-time on a wide range of commodity devices, and
(iii) is significantly more accurate than \GSplat{}.
We do not aim to outperform \ZipNeRF{} in terms of quality -- it is our model's teacher and represents an \emph{upper bound} on our method's achievable quality.

\begin{table}[!t]
	\centering
	\caption{
	    Results on the large scenes from \ZipNeRF{}~\cite{barron2023zipnerf}.
	    As the spatial subdivision resolution $\mathrm{K}$ increases, our accuracy approaches that of the \ZipNeRF{} ``teacher''.
	    \label{tab:zipnerf_quant}
	}
	\CaptionToTableVspace{}
	\resizebox{0.95\linewidth}{!}{
    	\Huge
		\begin{tabular}{@{\,}l|cccccc@{\,}}
			Method %
			    & PSNR$\uparrow$
			    & SSIM$\uparrow$
		        & LPIPS$\downarrow$
		        & FPS$\uparrow$
		        & Mem (MB)$\downarrow$
		        & Disk (MB)$\downarrow$
			\\
			\hline 
			MERF (ours)           &                      23.49 &                      0.746 &                      0.444 &  \cellcolor{tabthird}283 &  \cellcolor{tabthird}522 & \cellcolor{tabsecond}128 \\
3DGS                  &  \cellcolor{tabthird}25.50 &  \cellcolor{tabthird}0.810 &  \cellcolor{tabthird}0.369 &  \cellcolor{tabfirst}441 &  \cellcolor{tabfirst}227 &  \cellcolor{tabthird}212 \\
Ours ($\mathrm{K}=1$) &                      25.44 &                      0.777 &                      0.412 & \cellcolor{tabsecond}329 & \cellcolor{tabsecond}505 &  \cellcolor{tabfirst}118 \\
Ours ($\mathrm{K}=3$) & \cellcolor{tabsecond}27.09 & \cellcolor{tabsecond}0.823 & \cellcolor{tabsecond}0.350 &                      220 &                      1313 &                      1628 \\
Ours ($\mathrm{K}=5$) &  \cellcolor{tabfirst}27.28 &  \cellcolor{tabfirst}0.829 &  \cellcolor{tabfirst}0.339 &                      204 &                      1454 &                      4108
\\ 
\hline
Zip-NeRF & 27.37 & 0.836 & 0.305 & 0.25$^{*}$ &                    -       & 607			        
		\end{tabular}
	}
	\AfterTableVspace{}
	\caption{
	    Results on the \MipNeRFThreeSixty{} dataset~\cite{barron2022mipnerf360}.
	    \label{tab:mipnerf360_quant}
	}
	\CaptionToTableVspace{}
	\resizebox{\linewidth}{!}{
    	\Huge
		\begin{tabular}{l|cccccc}
			Method %
		        & PSNR$\uparrow$
    			    & SSIM$\uparrow$
			        & LPIPS$\downarrow$
    		        & FPS$\uparrow$
    			    & Mem (MB)$\downarrow$
			        & Disk (MB)$\downarrow$
			\\
			\hline 
            \BakedSDF{}              &                      24.51 &                      0.697 &                      0.309 &  \cellcolor{tabfirst}507 &                      573 &                      457 \\
\InstantNGP{}            &  \cellcolor{tabthird}25.68 &                      0.706 &  \cellcolor{tabthird}0.302 &                      8.61 &                    -       &  \cellcolor{tabfirst}104 \\
\GSplat{}                & \cellcolor{tabsecond}27.20 & \cellcolor{tabsecond}0.815 & \cellcolor{tabsecond}0.214 &  \cellcolor{tabthird}260 &                      780 &                      740 \\
\MERF{} (published)      &                      25.24 &                      0.722 &                      0.311 &                      162 &  \cellcolor{tabfirst}400 &                      162 \\
\MERF{} (ours)           &                      24.95 &  \cellcolor{tabthird}0.728 &  \cellcolor{tabthird}0.302 & \cellcolor{tabsecond}278 &  \cellcolor{tabthird}504 &  \cellcolor{tabthird}153 \\
\Ours{} ($\mathrm{K}=1$) &  \cellcolor{tabfirst}27.98 &  \cellcolor{tabfirst}0.818 &  \cellcolor{tabfirst}0.212 &                      217 & \cellcolor{tabsecond}466 & \cellcolor{tabsecond}139
\\ 
\hline
\ZipNeRF{} & 28.78 & 0.836 & 0.177 & 0.25$^{*}$ &                    -       & 607
        \end{tabular}
	}
	\AfterTableVspace{}
	\caption{
	    Frame rates for different devices and resolutions, averaged across the outdoor \MipNeRFThreeSixty{} scenes.
	    Our method achieves real-time rendering rates on all platforms, including smartphones.
	    See \appendixtableref{6}{tab:hardware} for hardware specs.
	    Not all methods are capable of running on all platforms.
    	\label{tab:framerates}
	}
	\CaptionToTableVspace{}
	\resizebox{0.775\linewidth}{!}{
    	\Huge
		\begin{tabular}{@{\,}l|cccc@{\,}}
			    & iPhone  %
			    & Macbook  %
			    & Desktop  %
			\\
			Method %
			    & $380\times640$
			    & $1280\times720$
			    & $1920\times1080$
			\\
			\hline 
        \BakedSDF{}              &                    -       &  \cellcolor{tabfirst}108 &  \cellcolor{tabfirst}412 \\
\GSplat{}                &                    -       &                    -       &  \cellcolor{tabthird}176 \\
\MERF{} (published)      &  \cellcolor{tabthird}38.1 &                      32 &                      113 \\
\MERF{} (ours)           &  \cellcolor{tabfirst}58.3 & \cellcolor{tabsecond}47.6 & \cellcolor{tabsecond}187 \\
\Ours{} ($\mathrm{K}=1$) & \cellcolor{tabsecond}55.4 &  \cellcolor{tabthird}42.5 &                      142
		\end{tabular}
	}
\end{table}

\subsection{Reconstruction Quality}
We first evaluate our method on four large scenes introduced by \ZipNeRF{}: \textsc{Berlin}, \textsc{Alameda}, \textsc{London}, and \textsc{Nyc}.
Each of these scenes was captured with 1,000-2,000 photos using a $180^{\circ}$ fisheye lens.
To produce an apples-to-applies comparison with \GSplat{}, we crop photos to a $110^{\circ}$ field-of-view and reestimate camera parameters with COLMAP~\cite{schoenberger2016sfm}.
Images are then downsampled to a resolution between $1392\times793$ and $2000\times1140$, depending on scene.
Because this is a recent dataset without established benchmark results, we limit evaluations to our method, \GSplat{}, \ZipNeRF{}, and \MERF{}.
As some scenes were captured with auto-exposure enabled, we modify our algorithm and all baselines accordingly (see \appendixref{A}{sec:model-details}).
\changed{We tuned 3DGS extensively for this dataset in correspondence with the original authors, see \appendixref{F}{sec:3dgs_attempted_improvements} for details.} 

The results shown in \cref{tab:zipnerf_quant} indicate that, for modest degrees of spatial subdivision $\SubmodelResolution{}$, the accuracy of our method strongly surpasses that of \MERF{} and \GSplat{}.
As $\SubmodelResolution{}$ increases, our model's reconstruction accuracy improves and approaches that of its \ZipNeRF{} teacher, with a gap of less than $0.1$ PSNR and $0.01$ SSIM at $\SubmodelResolution{} = 5$.

We find that these quantitative improvements understate the qualitative improvements in reconstruction accuracy, as demonstrated in \cref{fig:vs_3dgs}.
In large scenes, our method consistently models thin geometry, high-frequency textures, specular highlights, and far-away content where the real-time baselines fall short.
At the same time, we find that increasing submodel resolution naturally increases quality, particularly with respect to high-frequency textures: see \cref{fig:submodel_resolution}.
In practice, we find our renders to be virtually indistinguihable from \ZipNeRF{}, as shown in \cref{fig:vs_zipnerf}.

We further evaluate our method on the \MipNeRFThreeSixty{} dataset of unbounded indoor and outdoor scenes~\cite{barron2022mipnerf360}.
These scenes are smaller in volume than those in the \ZipNeRF{} dataset, and as such, spatial subdivision is unnecessary to achieve high quality results.
As shown in \cref{tab:mipnerf360_quant}, the $\SubmodelResolution{} = 1$ version of our model outperforms all prior real-time models on this benchmark in terms of image quality with a rendering speed comparable to that of \GSplat{}.
Note that we significantly improve upon the \MERF{} baseline despite the lack of spatial subdivision, which demonstrates the value of the our deferred appearance network partitioning and feature gating contributions.
Figures~\ref{fig:vs_merf} and \ref{fig:vs_zipnerf} illustrate this improvement: our method is better at representing high-frequency geometry and textures while eliminating distracting floaters and fog. Finally, in \cref{fig:submodel_consistency} you can observe how our ray jittering improves temporal stability of rendered views under camera motion.

\subsection{Rendering Speed}
We report frame rates at the native dataset resolution in Tables~\ref{tab:zipnerf_quant} and \ref{tab:mipnerf360_quant}.
We measure this by rendering each image in the test set $100 \times$ on a GPU-equipped workstation and computing the arithmetic mean of frame times (\ie, the harmonic mean of FPS).
See Appendices D and G for more benchmark details and results.

In \cref{tab:framerates}, we report the performance of several recent methods as a function of image resolution and hardware platform.
Our model is able to run in real-time on smartphones and laptops, albeit at reduced resolutions.
\MERF{} and \BakedSDF{}~\cite{bakedsdf} are also capable of running on resource-constrained platforms, but their reconstruction quality significantly lags behind ours; see Tables~\ref{tab:zipnerf_quant} and \ref{tab:mipnerf360_quant}.
We omit results for \BakedSDF{} on iPhone due to memory limitations and \GSplat{} on iPhone and MacBook, where an official implementation for non-CUDA devices is lacking.
While \GSplat{} modestly outperforms our cross-platform web viewer in terms of rendering speed on a desktop workstation, Tables~\ref{tab:zipnerf_quant} and \ref{tab:mipnerf360_quant} indicate that its reconstruction quality lags behind our method.

\moved{
\begin{table}[!t]
    \newcommand{\verticaltext}[2]{
      \parbox[t]{1mm}{\multirow{#1}{*}{\rotatebox[origin=c]{90}{#2}}} 
    }
	\centering
	\caption{
	    \moved{Ablation study of our model without subdivision ($\mathrm{K}=1$) on the \MipNeRFThreeSixty{} dataset~\cite{barron2022mipnerf360}.}
    	\label{tab:mipnerf360_ablations}
	}
	\CaptionToTableVspace{}
	\resizebox{0.9\linewidth}{!}{
    	\Huge
		\begin{tabular}{@{\,}ll|ccccc@{\,}}
			& %
		        & PSNR$\uparrow$
    			& SSIM$\uparrow$
			    & LPIPS$\downarrow$
			    & Disk (MB)$\downarrow$
			\\ \hline
            
\verticaltext{3}{Distill.} & No Color Supervision &                      26.45 &                      0.775 &                      0.259 &  \cellcolor{tabfirst}113 \\
 & No Geometry Supervision                        &                      27.90 &                      0.814 &                      0.218 &                      149 \\
 & No Ray Jittering                               &                      26.54 &                      0.776 &                      0.258 & \cellcolor{tabsecond}124 \\
\hline \verticaltext{4}{Optim.} & No SSIM Loss    &                      27.88 &                      0.814 &                      0.217 &                      142 \\
 & 25k Train Steps                                &                      27.56 &                      0.803 &                      0.231 &                      148 \\
 & No Larger Hash Grids                           &                      27.89 &  \cellcolor{tabthird}0.816 &                      0.215 &                      145 \\
 & No Hyperparam. Tuning                          &                      27.17 &                      0.785 &                      0.245 &                      148 \\
\hline \verticaltext{4}{Model} & No MLP Grid      &                      27.40 &                      0.811 &                      0.218 &                      148 \\
 & No Total Var. Reg.                             &  \cellcolor{tabthird}27.92 & \cellcolor{tabsecond}0.817 &  \cellcolor{tabthird}0.214 &                      143 \\
 & No Feature Gating                              &                      27.63 &                      0.809 &                      0.221 &  \cellcolor{tabthird}134 \\
 & No Median filter                               &  \cellcolor{tabfirst}28.00 &  \cellcolor{tabfirst}0.818 & \cellcolor{tabsecond}0.213 &                      153 \\
\hline & Ours                                     & \cellcolor{tabsecond}27.98 &  \cellcolor{tabfirst}0.818 &  \cellcolor{tabfirst}0.212 &                      139
        \end{tabular}
	}
\end{table}
\begin{table*}[t]
    \newcommand{\best}[1]{\cellcolor{red!40}#1}
    \newcommand{\secondbest}[1]{\cellcolor{orange!40}#1}
    \newcommand{\thirdbest}[1]{\cellcolor{yellow!40}#1}

	\small
	\centering

	\caption{
	    \moved{Spatial resolution.
	    We compare with monolithic variants of \MERF{} with distillation (\MERFDistill{}) and optimization (\MERFDistillTrain{}) contributions and varying triplane and sparse grid resolutions.
	    Rather than directly increasing resolution, our method instead increases the submodel subdivision $\SubmodelResolution{} = \{1 .. 5 \}$.
	    Our method achieves universally higher quality and greater spatial resolution while bounding memory usage by the size of the two largest submodels.
	    Note that this experiment uses a reduced version of our method with $2^{19}$ hash grid entries that was trained for only 50k steps.}
	}
	\CaptionToTableVspace{}
	\resizebox{0.75\linewidth}{!}{
    	\large
		\begin{tabular}{@{\,}l|c|cc|cc|cc|cc|cc@{\,}}
			& Spatial Res.
			    & \multicolumn{2}{c|}{$2048^3$}
			    & \multicolumn{2}{c|}{$4096^3$}
			    & \multicolumn{2}{c|}{$6144^3$}
			    & \multicolumn{2}{c}{$8192^3$}
			    & \multicolumn{2}{c}{$10240^3$}
			    \\
			& $\SubmodelResolution{}$
			    & PSNR$\uparrow$
			        & Mem$\downarrow$
			    & PSNR$\uparrow$
			        & Mem$\downarrow$
			    & PSNR$\uparrow$
			        & Mem$\downarrow$
			    & PSNR$\uparrow$
			        & Mem$\downarrow$
			    & PSNR$\uparrow$
			        & Mem$\downarrow$
			\\
			\hline 
        \MERF{} (ours) & 1      &                      23.14 & \cellcolor{tabsecond}433 &                      23.71 & \cellcolor{tabsecond}1974 &                      23.77 & \cellcolor{tabsecond}5590 &                      23.74 & \cellcolor{tabsecond}12119 &                    -       &                    -       \\
\MERFDistill{} & 1      &  \cellcolor{tabthird}24.14 &  \cellcolor{tabthird}460 &  \cellcolor{tabthird}24.96 &  \cellcolor{tabthird}2042 &  \cellcolor{tabthird}25.23 &  \cellcolor{tabthird}5745 &  \cellcolor{tabthird}25.28 &  \cellcolor{tabthird}12684 &                    -       &                    -       \\
\MERFDistillTrain{} & 1 & \cellcolor{tabsecond}24.43 &                      478 & \cellcolor{tabsecond}25.29 &                      2168 & \cellcolor{tabsecond}25.52 &                      6061 & \cellcolor{tabsecond}25.59 &                      13198 &                    -       &                    -       \\
\Ours{} & Varies        &  \cellcolor{tabfirst}25.28 &  \cellcolor{tabfirst}428 &  \cellcolor{tabfirst}26.36 &  \cellcolor{tabfirst}953 &  \cellcolor{tabfirst}26.85 &  \cellcolor{tabfirst}1067 &  \cellcolor{tabfirst}26.84 &  \cellcolor{tabfirst}1076 &  \cellcolor{tabfirst}27.06 &  \cellcolor{tabfirst}1156
		\end{tabular}
	}
	\label{tab:resolutions}
\end{table*}

\subsection{Feature Ablations}
In \cref{tab:mipnerf360_ablations}, we present an ablation study demonstrating the impact of the contributions introduced in this paper.
In particular, we examine the quantitative effect of independently disabling each feature on the \MipNeRFThreeSixty{} dataset.
We group contributions into three overarching themes: distillation training (\emph{Distill.}), optimization (\emph{Optim.}), and model architecture (\emph{Model}).
In the following, we describe each of these contributions in detail.

\emph{Distill.} considers the distillation training regime described in \cref{sec:distillation}.
In ``No Color Supervision'', we replace the teacher's color predictions with ground truth pixels.
As ground truth pixels do not support ray jittering, this feature is disabled as well.
``No Geometry Supervision'' replaces the teacher's proposed ray intervals with a learned proposal network as employed in \MERF{}.
The geometry reconstruction loss of \cref{eqn:geometry-reconstruction} is also disabled for this experiment.
In ``No Ray Jittering'', we disable ray jittering and restrict camera rays to those provided by the training set.

\emph{Optim.} concerns changes made to the capacity and training of the core \MERF{} architecture.
In ``No SSIM Loss'', we consider the effect of disabling the \dssim{} loss.
As patch-based inputs are no longer required, this ablation samples rays uniformly at random rather than as 3x3 patches.
In ``25k Train Steps'', we decrease the number of training steps from 200,000 to 25,000 to match \MERF{}.
In ``No Larger Hash Grids'', we reduce the number of entries in the multi-resolution hash encoding from $\NumHashGridEntries = 2^{22}$ to $2^{21}$.
We further decrease the depth of the MLP immediately following the hash encoding from 2 hidden layers to 1.
In ``No Hyperparam. Tuning'', we revert the various changes made to hyperparameters such as the hash encoding entry regularization weight, learning rate schedule, deferred network activation function, and a scaling factor applied to inputs of the contraction function (\cref{eqn:contract360}).

\emph{Model} ablates our contributions to the model architecture.
In ``No MLP Grid'', we eliminate deferred appearance network partitioning as introduced in \cref{sec:deferred-appearance-network-partitioning}.
In ``No Total Var. Regularization'', we disable the total variation regularization term encouraging similarity between spatially-adjacent deferred network parameters.
In ``No Feature Gating'', we eliminate feature gating, as described in \cref{sec:feature-gating}.
In ``No Median filter'', we omit the median filter applied to the high-resolution occupancy grid at baking time.

When removed one at a time, most contributions have a modest but non-negligible effect on reconstruction quality.
In isolation, we find ray jittering to be the single most valuable contribution to model quality.
The reason for this is a lack of supervision: training diverges on \textsc{treehill} when ray jittering is omitted.
We attribute this to the increased capacity of our model coupled with insufficient supervision.
In addition, we find hyperparameter tuning and a longer training schedule to be critical for the reconstruction of finer details, as demonstrated by a significant decrease in \ssim{}.
The elimination of a single contribution -- median filtering -- has a positive impact on quality.
This feature eliminates floaters and reduces baked asset size, but also introduces a slight discrepancy between the representation queried at training and rendering time.

\subsection{Spatial Resolution Ablation}
In \cref{tab:resolutions}, we investigate the effects of increased spatial resolution on reconstruction quality and representation size on a variety of \MERF{}-based baselines applied to the \ZipNeRF{} dataset.
While \OurMethod{}'s spatial resolution may be increased by introducing additional submodels, a natural alternative is increasing the resolution of baked assets directly.

To this effect, we consider three baselines: our implementation of \MERF{}, \MERF{} with distillation training (\MERFDistill{}), and \MERF{} with distillation training and model enhancements (\MERFDistillTrain{}).
For each baseline, we increase spatial resolution by multiplying triplane and sparse feature grid resolutions by a factor of 1, 2, 3, 4, and 5 from a base resolution of $ \TriplaneResolution = 2048$ and $\SparseGridResolution = 512$.
We compare to \OurMethod{}, where we instead multiply submodel resolution $\SubmodelResolution{}$ by the same amount, with each submodel retaining the base triplane and spares feature grid resolution.
To accelerate training, we employ a reduced version of our method in this experiment: models are trained for 50,000 steps and use a multi-resolution hash grid size of $\NumHashGridEntries = 2^{19}$ entries per hash table.

This experiment highlights the value of model subdivision.
Not only does our method consistently outperform the other model variants, memory usage of the real-time viewer grows slowly for choices of $\SubmodelResolution{} > 2$ and remains bounded by the size of a model's two largest submodels (See \cref{sec:rendering} for details).
In contrast, the memory requirements of the baselines continue to grow superlinearly with spatial resolution.

}

\subsection{Limitations}
While our method performs well in terms of quality and memory usage, it comes with a high storage cost.
In the live viewer, this results in loading events and high network usage.
Our method also incurs a non-trivial training cost: in addition to training the teacher, we optimize our method for 100,000-200,000 steps on 8x V100 or 16x A100 GPUs, depending on dataset.
\changed{End-to-end, our method requires approximately 33 hours on a representative \ZipNeRF{} scene (\textsc{Alameda}: 3 hours for teacher training, 24 for distillation, and 6 for baking) and 17 hours on a representative \MipNeRFThreeSixty{} scene (\textsc{Garden}: 2 hours for teacher training, 14 for distillation, and 1 for baking) with 16x A100 GPUs.}
While our method achieves higher quality than \GSplat{} on average, it is not universally higher in detail for all parts of all scenes.
We attribute this to the voxel structure imposed on the scene by our representation.

\section{Conclusion}
\label{sec:conclusion}

We present SMERF, a streamable, memory-efficent radiance field representation for real-time view-synthesis of large scenes.
Our method renders in real-time in the web browser on everyday, resource-constrained consumer devices including smartphones and laptops.
At the same time, it achieves higher quality than existing real-time methods on both medium and large scenes, exceeding the existing state-of-the-art by 0.78 and 1.78 dB PSNR, respectively.

We achieve this by distilling a high-fidelity \ZipNeRF{} teacher into a hierarchical student built on \MERF{}.
Our method subdivides scenes into independent submodels, each of which is further subdivided into a set of deferred rendering networks.
As a consequence, only a single submodel and a local neighborhood of deferred network parameters are required to render a target view. 
We further improve \MERF{}'s viewer, increasing frame rates by over 70\%.
As a result, memory and compute requirements remain on par with \MERF{} while markedly increasing quality and rendering speed.
For large scenes, our quality is nearly indistinguishable from \ZipNeRF{}, the current state-of-the-art in offline view-synthesis.
\IfAnonymous{}{
    We encourage readers to explore \OurMethod{} interactively on our project webpage: \OurWebsite{}.
}

\begin{acks}
\IfAnonymous{}{
    We would like to thank Dor Verbin, Pratul Srinivasan, Ben Mildenhall, Klaus Greff, Alexey Dosovitskiy, Marcos Seefelder, and in particular Stephan Garbin for discussions and valuable feedback throughout the project. 
    We would also like to sincerely thank Georgios Kopanas and Bernhard Kerbl for their generous help with tuning 3DGS and verifying our results.
    We would further like to thank Oliver Unke for his knowledge of differential geometry as applied to data augmentation and Ilies Ghanzouri for his contributions to the live renderer.
    Finally, we would like to thank Jane Kasdano for her valuable feedback on the design and presentation of this work.
}
\end{acks}

{
    \bibliographystyle{ACM-Reference-Format}
    \bibliography{main}
}

\clearpage
\appendix

\setcounter{equation}{14}
\setcounter{figure}{10}
\setcounter{table}{5}

\section{Model Details}
\label{sec:model-details}

In this section, we describe the design and architecture of \OurMethod{}.
We begin by describing the architecture employed in our experiments on the \MipNeRFThreeSixty{} dataset.
We then highlight the differences in experiments on the \ZipNeRF{} dataset.

\subsection{\MipNeRFThreeSixty{} Dataset}
In this dataset, a single submodel suffices to represent an entire scene.
Following \MERF{}, our submodel consists of (i) a multi-resolution hash encoding~\cite{muller2022instant} coupled with (ii) a feature-decoding MLP, followed by (iii) a deferred rendering network.
For (i), we use 20 levels of hash encoding, each with $\NumHashGridEntries = 2^{22}$ entries and $2$ features per entry.
Each level corresponds to a logarithmically-spaced resolution between $16^3$ and $8192^3$ voxels.
Upon look-up and concatention, we pass these features to a feature-decoding MLP with 2 hidden layer and 64 hidden units.
The MLP then produces 8 outputs corresponding to density, diffuse color, and feature preactivations.
Unlike \MERF{}, \OurMethod{} does not need to train Proposal-MLPs; instead, ray intervals are provided by the teacher.

After training, the outputs of (i) and (ii) are quantized and cached to form the triplane and sparse voxel representation used by the real-time renderer.
As in \MERF{}, we use a spatial resolution of $\TriplaneResolution^2 = 2048^2$ for triplanes and $\SparseGridResolution^3 = 512^3$ for sparse voxel grids (See \MainTextSectionRef{3}{sec:preliminaries}).

Our view-dependence model follows \SNeRG{}~\cite{hedman2021snerg}, i.e.\ we predict a view-dependent residual color as a post process after volume rendering each ray.
Instead of using a single view-dependence MLP, we trilinearly interpolate its parameters within a grid of $\DeferredResolution{}^3 = 5^3$ vertices (See \MainTextSectionRef{4}{sec:deferred-appearance-network-partitioning}). At each vertex, we assign parameters $\DeferredMlpParams_{uvw}$ for an MLP with 2 hidden layers and 16 hidden units.
These MLPs employ the exponential linear unit activation function~\cite{clevert2015fast}.

To focus our model on the center of each scene, we multiply position coordinates $\pposition$ by 2.5 before applying the contraction function in \MainTextEquationRef{5}{eqn:contract360}, \ie $\tilde{\pposition} = 2.5 \, \pposition$.
We find that this provides our method with significantly higher spatial fidelity in the central portion of the scene without any sharpness loss in the periphery.

\subsection{\ZipNeRF{} Dataset}
To account for the larger extent of the \ZipNeRF{} scenes, we fit up to $\SubmodelResolution^3 = 5^3$ submodels per scene.
As described in \MainTextSectionRef{4}{sec:coordinate-space-partitioning}, we instantiate submodel parameters only when a training camera lies within a subvolume's boundaries.
This results in fewer than 25 instantiated submodels per scene for all of our experiments.

The architecture of each individual submodel follows the \MipNeRFThreeSixty{} experiments with the following changes:
First, we reduce the number of hash encoding entries per level to $\NumHashGridEntries = 2^{21}$ in order to reduce memory consumption during training.
Second, we introduce a form of exposure conditioning inspired by \BlockNeRF{}~\cite{tancik2022blocknerf}.
Specifically, we replace the rendered feature inputs to the deferred rendering network as follows,
\begin{align}
    \hat{\RayFeature}(\Ray) &= \VolumetricRender\lft(
        \lft\{ \Density_k \rgt\},
        \lft\{ \PositionFeature(\PPosition_k) \rgt\}
    \rgt)\,, \\
    \RayFeature(\Ray) &= \Logit(\hat{\RayFeature}(\Ray)) + \log \left( \frac{\mathrm{ISO} \times \Delta t}{1000} \right)\,,
    \label{eq:ours_exposure}
\end{align}
where $\mathrm{ISO}$ is the input image's ISO speed rating and $\Delta t$ is the open shutter time in seconds as extracted from the training images' metadata.
In initial experiments, we fit models by instead fixing the exposure level of the teacher; however, we found visual quality lacking in scenes with large variations in illumination.

Finally, to avoid visible shearing at subvolume boundaries, we enlarge the Euclidean region of each submodel's contracted coordinate space.
Specifically, we scale the submodel position coordinates $\mathbf{x}$ by 0.8 before applying the contraction function in \MainTextEquationRef{5}{eqn:contract360}.

\section{Training}
\label{sec:student_training}

In this section, we describe the training of \OurMethod{}.
As in \appendixref{A}{sec:model-details}, we begin by describing training for \MipNeRFThreeSixty{} scenes, followed by \ZipNeRF{} scenes.

\subsection{\MipNeRFThreeSixty{} Scenes}
Unless otherwise stated, we optimize all models with Adam~\cite{kingma2015adam} for 200,000 steps on 16 A100 GPUs.
Our learning rate follows a cosine decay curve starting at 1e-2 and decaying to 3e-4 and our minibatches consist of 65,376 camera rays organized into 3x3 patches, where the teacher network provides 32 ray intervals per camera ray. We conserve memory by applying gradient accumulation, where each training step is accumulated over two subbatches of 32,688 camera rays.

Our models are optimized with respect to a weighted sum of the several loss functions. Photometric supervision is derived from two sources, as described in \MainTextEquationRef{10}{eqn:photometric-loss}: (i) a per-pixel photometric RMSE loss, which encourages rendered pixels to match the teacher's, with a weight of 1.0, and (ii) a DSSIM loss on 3x3 patches, which is added with a weight of 1.5.
To adapt the DSSIM loss to smaller patch sizes, we replace the 11x11 Gaussian blur kernel with a na\"ive mean.

For parameter regularization, we penalize the magnitude of hash encoding entries and the variation in deferred rendering network parameters.
In particular, we penalize the average squared magnitude of each hash encoding level's parameters $\HashGridParameters_{t} \in \Reals^{\NumHashGridEntries \times 2}$ using a weight of $\lossmult_{\HashGridRegularizer} = 0.01$, summing across different levels.
Within each submodel, we further penalize the squared difference between between MLP parameters of spatially-adjacent MLPs in the deferred rendering network with a weight of $\lossmult_{\DeferredNetworkRegularizer} = 0.1$.
\begin{align}
    \LossFn_{\HashGridRegularizer}
        &= \lossmult_{\HashGridRegularizer}
           \sum_{\HashGridIndex} \frac{1}{\NumHashGridEntries} \norm{ \HashGridParameters_{\HashGridIndex} }_2^2 \\
    \LossFn_{\DeferredNetworkRegularizer}
        &= \lossmult_{\DeferredNetworkRegularizer}
           \sum_{\Delta \in \{ 0, 1\}^3 } \frac{\norm{
               \DeferredMlpParams_{uvw} - \DeferredMlpParams_{uvw + \Delta}
           }_1}{|\DeferredMlpParams_{uvw}|}
\end{align}

The final loss function is thus,
\begin{align}
    \LossFn
        &= \LossFn_{\radiance} +  \LossFn_{\HashGridRegularizer}  + \LossFn_{\DeferredNetworkRegularizer}
    \label{eqn:final-loss-fn}
\end{align}

We further regularize our models through ray jittering and random submodel reassignment.
In particular, we independently perturb each camera ray by adding noise sampled from an isotropic Gaussian distribution with zero mean and $\sigma = 0.03 \SubmodelResolution{}$.
For each 3x3 patch, we further sample a random rotation matrix with $\epsilon = 0.03 \SubmodelResolution{}$.
Note that this rotation matrix is shared by all camera rays in the patch.
During training, we further reassign submodels to 20\% of camera rays to randomly-selected neighboring submodels (\ie
submodels whose centers are within $2$ units of the ray origin).

As in \MERF{}, we ignore ray segments with low opacity and volumetric rendering weights.
Specifically, we set the rendering weight to zero for segments whose contribution is below a threshold.
We initialize this threshold to 0 between steps 0 and 80,000, and then linearly increase it from 5e-4 to 5e-3 between steps 80,000 and 160,000.

\subsection{\ZipNeRF{} Scenes}
Unlike the \MipNeRFThreeSixty{} scenes, the \ZipNeRF{} scenes were captured with a $180^{\circ}$ field-of-view (FOV) fisheye lens.
For fair comparisons with methods that do not support fisheye lenses (\eg \GSplat{}), we crop these photos to a $110^{\circ}$ FOV and use COLMAP~\cite{schoenberger2016sfm} to estimate camera parameters and apply undistortion. 
The resulting undistorted photos have a maximum width of 4,000 pixels, which we downsample by $2 \times$ for training and evaluation.
This ``undistorted'' dataset is used for all quantitative results appearing in this work.
As neither \ZipNeRF{} nor \OurMethod{} are limited to undistorted cameras, we employ the original fisheye dataset for our rendered videos and demos.

Training for \ZipNeRF{} differs from \MipNeRFThreeSixty{} scenes as follows:
First, we introduce a submodel consistency loss with a weight of 1.0 as described in \MainTextEquationRef{14}{eqn:submodel-consistency-loss}, which is added to \MainTextEquationRef{19}{eqn:final-loss-fn}.
Second, to reduce memory usage, we accumulate gradients over 4 batches of 16,272 rays, resulting in a total batch size of 65,088 rays.
Finally, we train for 100,000 iterations, with the ray segment contribution threshold decaying between steps 40,000 and 80,000.
\section{Baking}
\label{sec:student_baking}

\subsection{\MipNeRFThreeSixty{} Scenes}
Our baking procedure is very similar to \MERF{}~\cite{reiser2023merf} with only minor differences.
We downsample our occupancy grid to $\SparseGridResolution{}^3 = 512^3$ with a maximum filter and then convert it to the ray marching acceleration distance grid used in \MainTextSectionRef{6.2}{sec:live-viewer}.

\subsection{\ZipNeRF{} Scenes}
In these scenes, we independently compute baked assets for each submodel.
To accelerate baking, we only use a subset of the training cameras to construct the high-resolution occupancy grids.
Recall that the world coordinate system is chosen such that cameras lie within a $[-\SubmodelResolution{}/2, \SubmodelResolution{}/2]^3$ cube.
To construct a submodel's occupancy grid, we only consider cameras whose origins lie within $1.5$ units of a submodel's center.
We further regularly subsample the camera rays by a factor of $2\times$ (\emph{i.e.} every second row and column) and independently apply ray jittering to each camera ray.
We find this process significantly reduces baking time with no visible impact on baked model quality.

\section{Real-time Rendering}
\label{sec:optimizations}
In addition to the modifications described in \MainTextSectionRef{6}{sec:live-viewer}, we introduce two optimizations that significantly accelerate rendering performance: a distance grid acceleration structure and a variety of low-level optimizations.
\subsection{Distance Grid}
\label{sec:distance-grid}
\changed{In addition to the optimizations described in \MainTextSectionRef{6.2}{sec:live-viewer}, we employ a \emph{distance grid} to accelerate ray marching in free space.}
While \MERF{} uses a hierarchy of binary occupancy grids to determine if a voxel is occupied or not, we use a single distance grid per submodel to store an 8-bit lower bound on the distance to the nearest occupied voxel.
These distance grids enable rapid skipping of large distances along camera rays without resorting to expensive hierarchical ray/box intersection tests.
\changed{We note that distance grids are directly derived from corresponding occupancy grids and that their use results in no additional approximations.}
As a result, our renderer achieves significantly higher frame rates with no impact on accuracy.

\subsection{Low-level Optimizations}
\label{sec:low-level-optimizations}
We further introduce two low-level improvements to the \MERF{} web viewer.
First, we compute the inputs to the Deferred Appearance MLP in a more efficient fashion --- rather than supporting a general mapping from channel index to input feature, we instead hard code the exact input features used by our model.
Second, we define the scene parameters (\eg grid resolutions and step sizes) as constants in the shader rather than uniform variables.
This aids the shader compiler in simplifying mathematical expressions.

\subsection{Frame Rate Benchmarking}
As the render time of most methods depend on the chosen viewpoints, we measure performance by re-rendering the test-set images in each scene.
This standardizes camera parameters, pose, and resolution.
For methods that rely on native low-level renderers~\cite{muller2022instant,kerbl20233d}, we instrument the code with GPU timers to measure the time taken to render each test-set image 100 times.
We were unable to render \ZipNeRF{} on any of our target platforms, so we estimate frame rate based on an 8x V100 GPU configuration.

Measuring performance accurately is more challenging for the methods that use browser renderers~\cite{yariv2023bakedsdf,reiser2023merf}, since web browsers provide lower precision CPU-only timers, limit the maximum frame rate to the update frequency of the screen, and often round the frame time to the nearest multiple of this frequency~\footnote{\url{https://bugs.webkit.org/show_bug.cgi?id=233312}}.
While disabling the frame rate limit of the browser is possible~\cite{hedman2021snerg}, we found this to be unreliable and prone to crashes.
To address this, we measure the time required to render each test set image 100 times while artificially limiting the frame rate to be lower than the update rate of the screen.
We achieve this by re-rendering the output image $k$ times before scheduling it for display.
Finally, to avoid reporting frame times that have been clamped to a multiple of the screen refresh rate, we measure performance with three different values for $k$ and record the smallest frame time for each test-set image.

In \MainTextTableRef{3}{tab:framerates}, we present results on three hardware platforms: iPhone, Macbook, and Desktop.
The exact specifications of these platforms are presented in \cref{tab:hardware}.
\begin{table}[h!]
	\small
	\centering
	\caption{
	    Hardware platforms used for frame rate evaluations.
	    Unlike other hardware platforms, iPhones use a shared CPU-GPU memory pool.
    	\label{tab:hardware}
	}
	\CaptionToTableVspace{}
	\resizebox{\linewidth}{!}{
    	\large
		\begin{tabular}{@{\,}l|cccc@{\,}}
		    Platform
    		    & Model
    		    & RAM
    		    & GPU
    		    & VRAM
    		    \\
    		\hline
		    iPhone
    		    & iPhone 15 Pro (2023)
    		    & 8 GB
    		    & Apple A17 Pro
    		    & -
    		    \\
		    MacBook
    		    & MacBook Pro (2019)
    		    & 32 GB
    		    & AMD Radeon Pro 5500M 
    		    & 8 GB
    		    \\
		    Desktop
    		    & ThinkStation P620 (2021)
    		    & 256 GB
    		    & NVIDIA RTX 3090
    		    & 24 GB
    		    \\
		\end{tabular}
	}
\end{table}

\section{\ZipNeRF{} Teacher}
\label{sec:teacher_training}

To train \OurMethod{}, we employ a variation of \ZipNeRF{}~\cite{barron2023zipnerf} as a teacher.
Unless otherwise stated, our hyperparameters match that of the original work.

\subsection{\MipNeRFThreeSixty{} scenes}
To improve quality, we make a small number of hyperparameter changes.
Our teacher models employ 10 hash grid levels with $\NumHashGridEntries = 2^{22}$ entries each, rather than the $2^{21}$ of the original work.
We train these for twice as long (50,000 steps) with an exponentially decaying learning rate between $10^{-2}$ and $10^{-3}$.
We also ramp up learning rate over 2,500, rather than 5,000, steps.
Finally, to reduce floater artifacts, we incorporate Gradient Scaling~\cite{philip2023gradient} with $\sigma = 0.3$.

\subsection{\ZipNeRF{} scenes}
We make further adjustments to account for these larger scenes.
In particular, we increase spatial resolution by adding an 11th hash grid level with a resolution of 16,384.
We also train for 100,000 steps, with a final learning rate of $10^{-4}$.

One of the \ZipNeRF{} scenes (\textsc{Alameda}) was captured using varying levels of exposure.
To account for this, we add \MainTextEquationRef{20}{eq:exposure} to the bottleneck features passed to the view-dependent MLP, as proposed in \BlockNeRF{}~\cite{tancik2022blocknerf}.

For the rendered videos and live demos, we further employ Affine Generative Latent Optimization~\cite{barron2023zipnerf}, which reduces the fog and discoloration introduced by phenomena such as photographer shadows and subtle changes in weather.
As the evaluation procedure for NeRF models employing GLO is still an open problem, we do not use GLO for any of the quantitative results.

\begin{figure}
    \centering
    \includegraphics[width=0.85\linewidth]{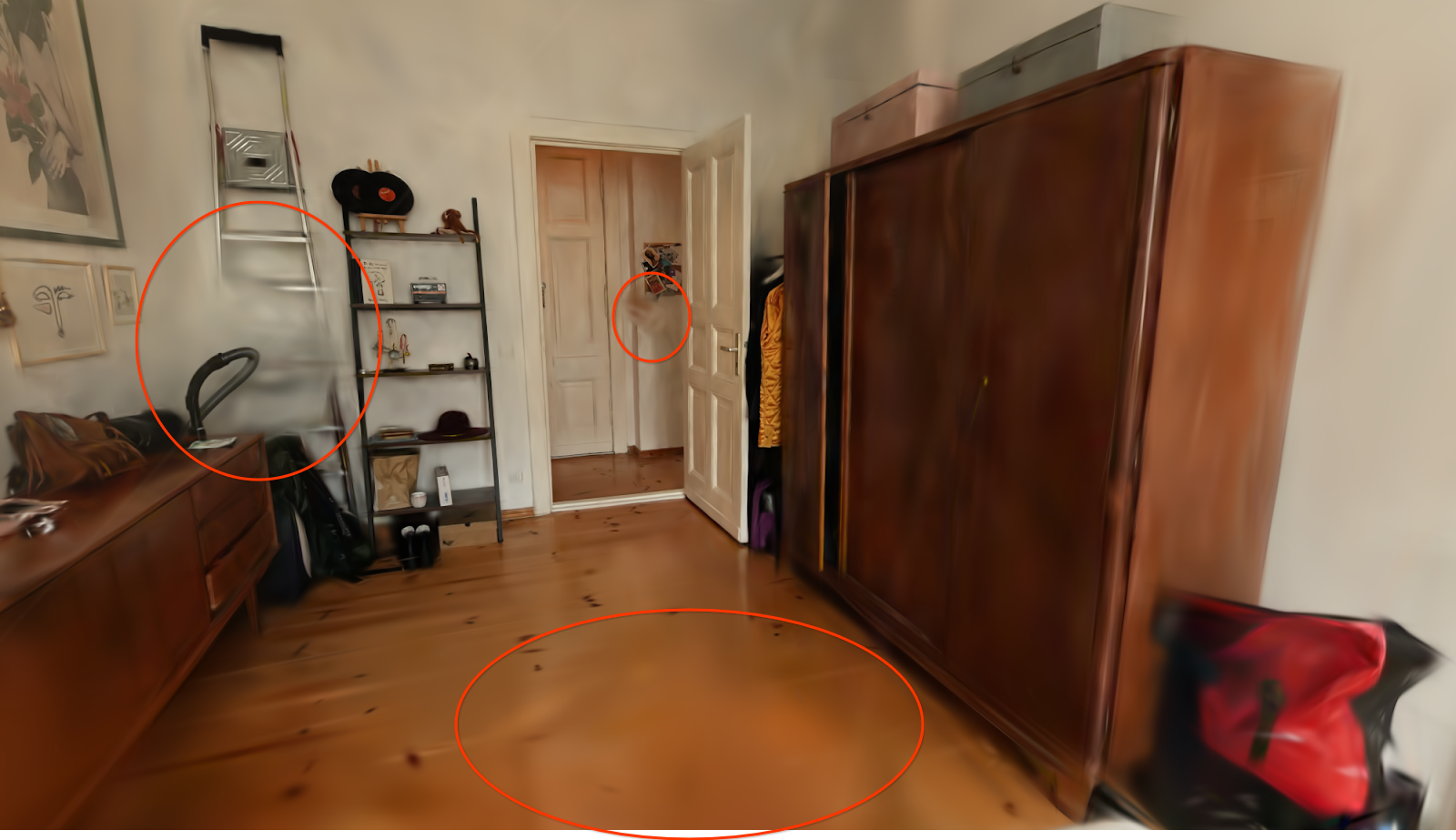}
    \caption{
        3D Gaussian Splatting fails to allocate a sufficient density of Gaussian primitives in under-supervised parts of the scene.
    }
    \label{fig:3dgs_blobs}
    \vspace{-0.125in}
\end{figure}

\section{3D Gaussian Splatting Experiments}
\label{sec:appendix_3dgs}
In this section, we elaborate on experiments and modifications made to 3D Gaussian Splatting (\GSplat{}) in this work.

\subsection{Hyperparameter Tuning}
\label{sec:3dgs_attempted_improvements}
As demonstrated in \cref{fig:3dgs_blobs}, \GSplat{} frequently struggles to allocate a sufficient density of Gaussian primitives throughout larger \ZipNeRF{} scenes.
While we find traces of this same issue in \MipNeRFThreeSixty{} scenes (See \MainTextFigureRef{5}{fig:vs_3dgs}), the limitation become particular clear in larger captures.
We believe this to be a limitation of the densification routine rather than the representation itself.
Unintuitively, we find \GSplat{} representations for \ZipNeRF{} scenes to contain, on average, 3.1x fewer primitives than \MipNeRFThreeSixty{} scenes, see \changed{Tables 1 and 2}.

Over the course of this work, we attempted to tune a wide range of hyperparameters to fix this issue.
We found that varying the number of training steps, densification steps, densification gradient thresholds, and learning rates universally reduces reconstruction quality compared to the default hyperparamters.
Lowering the densification threshold increases the number of primitives as well as number of floating artifacts.
In correspondences with the original authors, they affirmed that our investigations went ``above and beyond what is considered best effort for comparisons.''

\subsection{Exposure}
\label{sec:exposure3dgs}
To fairly compare \GSplat{} on the \ZipNeRF{} dataset, we extended the method to explicitly capture variation in camera exposure.
We found that this significantly improves quality on scenes where exposure varies (\textsc{Alameda}).
We modified \GSplat{} as follows:

Similarly to \cref{eq:ours_exposure}, we define exposure of image $i$ as,
\begin{equation}
    \Exposure_i = \log\lft(\frac{\ISO_i \times \Delta t_i}{1000}\rgt) \, ,
    \label{eq:exposure}
\end{equation}
and then feed the color predicted by \GSplat{} and the exposure value $\Exposure_i$ to an MLP with 4 hidden layers and 10 units.
The output of the MLP is the exposure-compensated RGB color.
The MLP is optimized in the main training loop with the Adam optimizer with learning rate of $10^{-3}$.

We initialize MLP parameters via ``pre-optimization'', where we bias the network towards the identity function by randomly sampling colors from the COLMAP point cloud and $\Exposure_i$ values from the training cameras.
Given a sampled color and exposure value, the network is optimized to predict the input color.
We run pre-optimization for 1,000 steps with a batch size of $2^{16}$ pixels.

For evaluation, we feed the ground-truth $\Exposure_i$ value to the model to obtain exposure-appropriate predictions.
When rendering output videos, we use the median $\Exposure_i$ value across the dataset.
\section{Additional Results}
\label{sec:additional-results}
We report PSNR, SSIM, LPIPS as well as video memory, disk usage, and frames per second for each scene in the \MipNeRFThreeSixty{} and \ZipNeRF{} datasets.
As in prior work~\cite{barron2022mipnerf360,kerbl20233d}, we compute LPIPS results using the implementation from \citet{zhang2018perceptual} with the VGG back-bone and provide it with RGB pixel intensities in $[0, 1]^3$.
We benchmark frames per second results at the native test-set resolution on the ``Desktop'' platform descrbed in \cref{tab:hardware}.

\begin{table*}%
\newcommand{\MySubTable}[2]{
	\centering
	\caption{#2}
	\vspace{-0.125in}
	\resizebox{0.8\linewidth}{!}{#1}
	\vspace{0.200in}
}
\begin{minipage}{\columnwidth}
    \MySubTable{		\begin{tabular}{@{\,}l|cccc@{\,}}
 & \textsc{berlin} & \textsc{nyc} & \textsc{alameda} & \textsc{london} \\
\hline
\GSplat{}                           &                      26.60 &                      26.41 &                      23.52 &                      25.45 \\
\MERF{} (ours)                      &                      25.27 &                      24.82 &                      20.34 &                      23.53 \\
\Ours{} ($\SubmodelResolution = 1$) &                      26.79 &                      25.40 &                      23.71 &                      25.86 \\
\Ours{} ($\SubmodelResolution = 3$) &  \cellcolor{tabthird}28.41 &  \cellcolor{tabthird}27.88 &  \cellcolor{tabthird}25.17 &  \cellcolor{tabthird}26.90 \\
\Ours{} ($\SubmodelResolution = 5$) & \cellcolor{tabsecond}28.52 & \cellcolor{tabsecond}28.21 & \cellcolor{tabsecond}25.35 & \cellcolor{tabsecond}27.05 \\
\hline
\ZipNeRF{}                          &  \cellcolor{tabfirst}28.59 &  \cellcolor{tabfirst}28.42 &  \cellcolor{tabfirst}25.41 &  \cellcolor{tabfirst}27.06
		\end{tabular}}{PSNR on \ZipNeRF{} scenes.}
    \MySubTable{		\begin{tabular}{@{\,}l|cccc@{\,}}
 & \textsc{berlin} & \textsc{nyc} & \textsc{alameda} & \textsc{london} \\
\hline
\GSplat{}                           &                      0.879 &                      0.829 &                      0.733 &                      0.797 \\
\MERF{} (ours)                      &                      0.840 &                      0.765 &                      0.646 &                      0.737 \\
\Ours{} ($\SubmodelResolution = 1$) &                      0.852 &                      0.771 &                      0.701 &                      0.789 \\
\Ours{} ($\SubmodelResolution = 3$) &  \cellcolor{tabthird}0.884 &  \cellcolor{tabthird}0.834 &  \cellcolor{tabthird}0.752 &  \cellcolor{tabthird}0.825 \\
\Ours{} ($\SubmodelResolution = 5$) & \cellcolor{tabsecond}0.887 & \cellcolor{tabsecond}0.844 & \cellcolor{tabsecond}0.758 & \cellcolor{tabsecond}0.829 \\
\hline
\ZipNeRF{}                          &  \cellcolor{tabfirst}0.891 &  \cellcolor{tabfirst}0.850 &  \cellcolor{tabfirst}0.767 &  \cellcolor{tabfirst}0.835
		\end{tabular}
}{SSIM on \ZipNeRF{} scenes.}
    \MySubTable{		\begin{tabular}{@{\,}l|cccc@{\,}}
 & \textsc{berlin} & \textsc{nyc} & \textsc{alameda} & \textsc{london} \\
\hline
\GSplat{}                           &                      0.335 &                      0.343 &                      0.407 &                      0.392 \\
\MERF{} (ours)                      &                      0.395 &                      0.426 &                      0.501 &                      0.456 \\
\Ours{} ($\SubmodelResolution = 1$) &                      0.380 &                      0.417 &                      0.445 &                      0.407 \\
\Ours{} ($\SubmodelResolution = 3$) &  \cellcolor{tabthird}0.330 &  \cellcolor{tabthird}0.336 &  \cellcolor{tabthird}0.381 &  \cellcolor{tabthird}0.352 \\
\Ours{} ($\SubmodelResolution = 5$) & \cellcolor{tabsecond}0.325 & \cellcolor{tabsecond}0.321 & \cellcolor{tabsecond}0.370 & \cellcolor{tabsecond}0.342 \\
\hline
\ZipNeRF{}                          &  \cellcolor{tabfirst}0.297 &  \cellcolor{tabfirst}0.281 &  \cellcolor{tabfirst}0.338 &  \cellcolor{tabfirst}0.304
		\end{tabular}
}{LPIPS on \ZipNeRF{} scenes.}
\end{minipage}
\hfill  %
\begin{minipage}{\columnwidth}
    \MySubTable{		\begin{tabular}{@{\,}l|cccc@{\,}}
 & \textsc{berlin} & \textsc{nyc} & \textsc{alameda} & \textsc{london} \\
\hline
\GSplat{}                           &  \cellcolor{tabfirst}215 &  \cellcolor{tabfirst}270 &  \cellcolor{tabfirst}250 &  \cellcolor{tabfirst}168 \\
\MERF{} (ours)                      &  \cellcolor{tabthird}504 &  \cellcolor{tabthird}466 &  \cellcolor{tabthird}546 & \cellcolor{tabsecond}571 \\
\Ours{} ($\SubmodelResolution = 1$) & \cellcolor{tabsecond}472 & \cellcolor{tabsecond}449 & \cellcolor{tabsecond}511 &  \cellcolor{tabthird}590 \\
\Ours{} ($\SubmodelResolution = 3$) &                      1324 &                      1169 &                      1303 &                      1456 \\
\Ours{} ($\SubmodelResolution = 5$) &                      1503 &                      1400 &                      1445 &                      1469 \\
\hline
\ZipNeRF{}                          &                    -       &                    -       &                    -       &                    -      
		\end{tabular}
}{VRAM usage on \ZipNeRF{} scenes. Measured in MB.}
    \MySubTable{		\begin{tabular}{@{\,}l|cccc@{\,}}
 & \textsc{berlin} & \textsc{nyc} & \textsc{alameda} & \textsc{london} \\
\hline
\GSplat{}                           &  \cellcolor{tabthird}202 &  \cellcolor{tabthird}258 &  \cellcolor{tabthird}239 &  \cellcolor{tabfirst}158 \\
\MERF{} (ours)                      & \cellcolor{tabsecond}105 & \cellcolor{tabsecond}61 & \cellcolor{tabsecond}160 & \cellcolor{tabsecond}187 \\
\Ours{} ($\SubmodelResolution = 1$) &  \cellcolor{tabfirst}72 &  \cellcolor{tabfirst}47 &  \cellcolor{tabfirst}129 &  \cellcolor{tabthird}226 \\
\Ours{} ($\SubmodelResolution = 3$) &                      1513 &                      925 &                      1802 &                      2272 \\
\Ours{} ($\SubmodelResolution = 5$) &                      4175 &                      3044 &                      4675 &                      4538 \\
\hline
\ZipNeRF{}                          &                      662 &                      662 &                      662 &                      662
		\end{tabular}
}{Disk usage on \ZipNeRF{} scenes. Measured in MB.}
    \MySubTable{ 		\begin{tabular}{@{\,}l|cccc@{\,}}
 & \textsc{berlin} & \textsc{nyc} & \textsc{alameda} & \textsc{london} \\
\hline
\GSplat{}                           &  \cellcolor{tabfirst}339.0 &  \cellcolor{tabthird}416.7 &  \cellcolor{tabfirst}467.3 &  \cellcolor{tabfirst}625.0 \\
\MERF{} (ours)                      &  \cellcolor{tabthird}184.8 & \cellcolor{tabsecond}469.5 &  \cellcolor{tabthird}269.5 &  \cellcolor{tabthird}348.4 \\
\Ours{} ($\SubmodelResolution = 1$) & \cellcolor{tabsecond}220.3 &  \cellcolor{tabfirst}480.8 & \cellcolor{tabsecond}358.4 & \cellcolor{tabsecond}363.6 \\
\Ours{} ($\SubmodelResolution = 3$) &                      140.8 &                      314.5 &                      236.4 &                      273.2 \\
\Ours{} ($\SubmodelResolution = 5$) &                      131.4 &                      261.8 &                      229.9 &                      260.4 \\
\hline
\ZipNeRF{}                          &                    0.25*       &                    0.25*       &                    0.25*      &                    0.25*      
		\end{tabular}}{FPS on \ZipNeRF{} scenes.}
\end{minipage}
\label{tab:scenes-zipnerf}
\end{table*}

\begin{table*}%
	\centering
	\caption{
	    PSNR on \MipNeRFThreeSixty{} scenes.
	}
	\CaptionToTableVspace{}
	\resizebox{0.85\linewidth}{!}{
		\begin{tabular}{@{\,}l|ccccc|cccc@{\,}}
 & \textsc{bicycle} & \textsc{flowers} & \textsc{garden} & \textsc{stump} & \textsc{treehill} & \textsc{room} & \textsc{counter} & \textsc{kitchen} & \textsc{bonsai} \\
 \hline
\BakedSDF{}                         &                      22.04 &                      19.53 &                      24.94 &                      23.59 &                      22.25 &                      28.68 &                      25.69 &                      26.72 &                      27.17 \\
\InstantNGP{}                       &                      22.79 &                      19.19 &                      25.26 &                      24.80 &                      22.46 &                      30.31 &                      26.21 &                      29.00 &                      31.08 \\
\GSplat{}                           &  \cellcolor{tabthird}25.25 &                      21.52 &  \cellcolor{tabthird}27.41 &  \cellcolor{tabthird}26.55 &                      22.49 &  \cellcolor{tabthird}30.63 &  \cellcolor{tabthird}28.70 &                      30.32 &  \cellcolor{tabthird}31.98 \\
\MERF{} (published)                 &                      22.62 &                      20.33 &                      25.58 &                      25.04 &                      22.39 &                      29.28 &                      25.82 &                      27.42 &                      28.68 \\
\MERF{} (ours)                      &                      22.44 &                      20.36 &                      25.89 &                      25.23 &                      22.37 &                      27.05 &                      24.64 &                      28.22 &                      28.40 \\
\MERFDistill{}                      &                      23.29 &                      20.83 &                      26.23 &                      25.78 &                      23.32 &                      29.87 &                      26.96 &                      28.63 &                      28.69 \\
\MERFDistillTrain{}                 &                      25.04 &  \cellcolor{tabthird}21.89 &                      26.77 &                      26.52 &  \cellcolor{tabthird}23.85 &                      30.58 &                      27.70 &  \cellcolor{tabthird}30.59 &                      30.89 \\
\Ours{} ($\SubmodelResolution = 1$) & \cellcolor{tabsecond}25.58 & \cellcolor{tabsecond}22.24 & \cellcolor{tabsecond}27.66 & \cellcolor{tabsecond}27.19 &  \cellcolor{tabfirst}23.93 & \cellcolor{tabsecond}31.38 & \cellcolor{tabsecond}29.02 & \cellcolor{tabsecond}31.68 & \cellcolor{tabsecond}33.19 \\
 \hline
\ZipNeRF{}                          &  \cellcolor{tabfirst}25.91 &  \cellcolor{tabfirst}22.46 &  \cellcolor{tabfirst}28.50 &  \cellcolor{tabfirst}27.63 & \cellcolor{tabsecond}23.88 &  \cellcolor{tabfirst}32.71 &  \cellcolor{tabfirst}30.10 &  \cellcolor{tabfirst}32.68 &  \cellcolor{tabfirst}35.12
		\end{tabular}
	}
	\label{tab:scenes-mipnerf360-psnr}
\end{table*}
\begin{table*}%
	\centering
	\caption{
	    SSIM on mip-NeRF 360 scenes.
	}
	\CaptionToTableVspace{}
	\resizebox{0.85\linewidth}{!}{
		\begin{tabular}{@{\,}l|ccccc|cccc@{\,}}
 & \textsc{bicycle} & \textsc{flowers} & \textsc{garden} & \textsc{stump} & \textsc{treehill} & \textsc{room} & \textsc{counter} & \textsc{kitchen} & \textsc{bonsai} \\
 \hline
\BakedSDF{}                         &                      0.570 &                      0.452 &                      0.751 &                      0.595 &                      0.559 &                      0.870 &                      0.808 &                      0.817 &                      0.851 \\
\InstantNGP{}                       &                      0.540 &                      0.378 &                      0.709 &                      0.654 &                      0.547 &                      0.893 &                      0.845 &                      0.857 &                      0.924 \\
\GSplat{}                           & \cellcolor{tabsecond}0.771 &                      0.605 & \cellcolor{tabsecond}0.868 &  \cellcolor{tabthird}0.775 &                      0.638 &  \cellcolor{tabthird}0.914 & \cellcolor{tabsecond}0.905 & \cellcolor{tabsecond}0.922 &  \cellcolor{tabthird}0.938 \\
\MERF{} (published)                 &                      0.595 &                      0.492 &                      0.763 &                      0.677 &                      0.554 &                      0.874 &                      0.819 &                      0.842 &                      0.884 \\
\MERF{} (ours)                      &                      0.609 &                      0.500 &                      0.774 &                      0.688 &                      0.570 &                      0.867 &                      0.815 &                      0.850 &                      0.883 \\
\MERFDistill{}                      &                      0.633 &                      0.521 &                      0.791 &                      0.713 &                      0.594 &                      0.888 &                      0.836 &                      0.857 &                      0.891 \\
\MERFDistillTrain{}                 &                      0.749 &  \cellcolor{tabthird}0.609 &                      0.831 &                      0.768 &  \cellcolor{tabthird}0.673 &                      0.909 &                      0.867 &                      0.906 &                      0.926 \\
\Ours{} ($\SubmodelResolution = 1$) &  \cellcolor{tabthird}0.760 & \cellcolor{tabsecond}0.626 &  \cellcolor{tabthird}0.844 & \cellcolor{tabsecond}0.784 & \cellcolor{tabsecond}0.682 & \cellcolor{tabsecond}0.918 &  \cellcolor{tabthird}0.892 &  \cellcolor{tabthird}0.916 & \cellcolor{tabsecond}0.941 \\
 \hline
\ZipNeRF{}                          &  \cellcolor{tabfirst}0.784 &  \cellcolor{tabfirst}0.655 &  \cellcolor{tabfirst}0.872 &  \cellcolor{tabfirst}0.807 &  \cellcolor{tabfirst}0.688 &  \cellcolor{tabfirst}0.927 &  \cellcolor{tabfirst}0.909 &  \cellcolor{tabfirst}0.926 &  \cellcolor{tabfirst}0.953
		\end{tabular}
	}
	\label{tab:scenes-mipnerf360-ssim}
\end{table*}
\begin{table*}%
	\centering
	\caption{
	    LPIPS on mip-NeRF 360 scenes.
	}
	\CaptionToTableVspace{}
	\resizebox{0.85\linewidth}{!}{
		\begin{tabular}{@{\,}l|ccccc|cccc@{\,}}
 & \textsc{bicycle} & \textsc{flowers} & \textsc{garden} & \textsc{stump} & \textsc{treehill} & \textsc{room} & \textsc{counter} & \textsc{kitchen} & \textsc{bonsai} \\
 \hline
\BakedSDF{}                         &                      0.368 &                      0.429 &                      0.213 &                      0.371 &                      0.366 &                      0.251 &                      0.286 &                      0.237 &                      0.259 \\
\InstantNGP{}                       &                      0.398 &                      0.441 &                      0.255 &                      0.339 &                      0.420 &                      0.242 &                      0.255 &                      0.170 &  \cellcolor{tabthird}0.198 \\
\GSplat{}                           & \cellcolor{tabsecond}0.201 &                      0.336 &  \cellcolor{tabfirst}0.103 & \cellcolor{tabsecond}0.210 &                      0.317 &                      0.221 & \cellcolor{tabsecond}0.204 & \cellcolor{tabsecond}0.130 &                      0.206 \\
\MERF{} (published)                 &                      0.371 &                      0.406 &                      0.215 &                      0.309 &                      0.414 &                      0.292 &                      0.307 &                      0.224 &                      0.262 \\
\MERF{} (ours)                      &                      0.356 &                      0.396 &                      0.206 &                      0.299 &                      0.386 &                      0.297 &                      0.312 &                      0.209 &                      0.259 \\
\MERFDistill{}                      &                      0.334 &                      0.388 &                      0.191 &                      0.284 &                      0.361 &                      0.264 &                      0.286 &                      0.204 &                      0.250 \\
\MERFDistillTrain{}                 &                      0.231 &  \cellcolor{tabthird}0.317 &                      0.149 &                      0.234 &  \cellcolor{tabthird}0.275 &  \cellcolor{tabthird}0.217 &                      0.243 &                      0.147 &                      0.200 \\
\Ours{} ($\SubmodelResolution = 1$) &  \cellcolor{tabthird}0.225 & \cellcolor{tabsecond}0.305 &  \cellcolor{tabthird}0.141 &  \cellcolor{tabthird}0.220 & \cellcolor{tabsecond}0.266 & \cellcolor{tabsecond}0.208 &  \cellcolor{tabthird}0.212 &  \cellcolor{tabthird}0.134 & \cellcolor{tabsecond}0.191 \\
 \hline
\ZipNeRF{}                          &  \cellcolor{tabfirst}0.188 &  \cellcolor{tabfirst}0.254 & \cellcolor{tabsecond}0.104 &  \cellcolor{tabfirst}0.181 &  \cellcolor{tabfirst}0.221 &  \cellcolor{tabfirst}0.189 &  \cellcolor{tabfirst}0.173 &  \cellcolor{tabfirst}0.117 &  \cellcolor{tabfirst}0.167
		\end{tabular}
	}
	\label{tab:scenes-mipnerf360-lpips}
\end{table*}
\begin{table*}%
	\centering
	\caption{
	    VRAM on mip-NeRF 360 scenes. Measured in MB.
	}
	\CaptionToTableVspace{}
	\resizebox{0.85\linewidth}{!}{
		\begin{tabular}{@{\,}l|ccccc|cccc@{\,}}
 & \textsc{bicycle} & \textsc{flowers} & \textsc{garden} & \textsc{stump} & \textsc{treehill} & \textsc{room} & \textsc{counter} & \textsc{kitchen} & \textsc{bonsai} \\
\hline
\BakedSDF{}                         &                      784 &                      879 & \cellcolor{tabsecond}369 &                      681 &                      796 & \cellcolor{tabsecond}287 &  \cellcolor{tabthird}423 &                      475 &                      460 \\
\InstantNGP{}                       &                    -       &                    -       &                    -       &                    -       &                    -       &                    -       &                    -       &                    -       &                    -       \\
\GSplat{}                           &                      1398 &                      829 &                      1330 &                      1131 &                      863 &  \cellcolor{tabthird}363 & \cellcolor{tabsecond}279 & \cellcolor{tabsecond}422 &  \cellcolor{tabfirst}284 \\
\MERF{} (published)                 &  \cellcolor{tabfirst}291 &  \cellcolor{tabfirst}271 &  \cellcolor{tabfirst}198 &  \cellcolor{tabfirst}228 &                      1630 &  \cellcolor{tabfirst}197 &  \cellcolor{tabfirst}254 &  \cellcolor{tabfirst}233 & \cellcolor{tabsecond}295 \\
\MERF{} (ours)                      &                      539 &                      537 &                      443 &                      478 &  \cellcolor{tabthird}561 &                      459 &                      511 &                      477 &                      546 \\
\MERFDistill{}                      &                      554 &                      536 &                      441 &                      486 &                      575 &                      412 &                      475 &                      443 &                      484 \\
\MERFDistillTrain{}                 &  \cellcolor{tabthird}508 &  \cellcolor{tabthird}515 &                      435 &  \cellcolor{tabthird}475 & \cellcolor{tabsecond}546 &                      469 &                      485 &                      445 &                      448 \\
\Ours{} ($\SubmodelResolution = 1$) & \cellcolor{tabsecond}502 & \cellcolor{tabsecond}500 &  \cellcolor{tabthird}431 & \cellcolor{tabsecond}463 &  \cellcolor{tabfirst}530 &                      464 &                      467 &  \cellcolor{tabthird}434 &  \cellcolor{tabthird}439 \\
\hline
\ZipNeRF{}                          &                    -       &                    -       &                    -       &                    -       &                    -       &                    -       &                    -       &                    -       &                    -      
		\end{tabular}
	}
	\label{tab:scenes-mipnerf360-memory}
\end{table*}
\begin{table*}%
	\centering
	\caption{
	    Disk usage on mip-NeRF 360 scenes. Sizes in MB.
	}
	\CaptionToTableVspace{}
	\resizebox{0.85\linewidth}{!}{
		\begin{tabular}{@{\,}l|ccccc|cccc@{\,}}
 & \textsc{bicycle} & \textsc{flowers} & \textsc{garden} & \textsc{stump} & \textsc{treehill} & \textsc{room} & \textsc{counter} & \textsc{kitchen} & \textsc{bonsai} \\
 \hline
\BakedSDF{}                         &                      932 &                      1039 &                      443 &                      808 &                      955 &                      296 &                      502 &                      568 &                      560 \\
\InstantNGP{}                       &  \cellcolor{tabfirst}106 &  \cellcolor{tabfirst}106 &  \cellcolor{tabthird}104 &  \cellcolor{tabfirst}105 &  \cellcolor{tabfirst}105 &                      103 &  \cellcolor{tabfirst}103 & \cellcolor{tabsecond}103 &  \cellcolor{tabthird}103 \\
\GSplat{}                           &                      1352 &                      809 &                      1289 &                      1108 &                      828 &                      329 &                      261 &                      406 &                      260 \\
\MERF{} (published)                 &                      207 &                      189 &                      114 &                      273 & \cellcolor{tabsecond}188 & \cellcolor{tabsecond}92 &                      136 &                      118 &                      167 \\
\MERF{} (ours)                      &                      197 &                      197 &  \cellcolor{tabthird}104 &  \cellcolor{tabthird}144 &  \cellcolor{tabthird}216 &  \cellcolor{tabthird}99 &                      145 &                      120 &                      189 \\
\MERFDistill{}                      &                      206 &                      188 & \cellcolor{tabsecond}102 &                      153 &                      236 &  \cellcolor{tabfirst}57 & \cellcolor{tabsecond}119 &  \cellcolor{tabfirst}96 &                      134 \\
\MERFDistillTrain{}                 & \cellcolor{tabsecond}187 &  \cellcolor{tabthird}187 & \cellcolor{tabsecond}102 &                      148 &                      217 &                      123 &  \cellcolor{tabthird}127 &                      107 &  \cellcolor{tabfirst}96 \\
\Ours{} ($\SubmodelResolution = 1$) &  \cellcolor{tabthird}190 & \cellcolor{tabsecond}184 &  \cellcolor{tabfirst}101 & \cellcolor{tabsecond}142 &                      217 &                      132 &                      130 &  \cellcolor{tabthird}104 & \cellcolor{tabsecond}98 \\
 \hline
\ZipNeRF{}                          &                      562 &                      562 &                      562 &                      562 &                      562 &                      562 &                      562 &                      562 &                      562

		\end{tabular}
	}
	\label{tab:scenes-mipnerf360-disk}
\end{table*}
\begin{table*}%
	\centering
	\caption{
	    FPS on mip-NeRF 360 scenes.
	}
	\CaptionToTableVspace{}
	\resizebox{0.85\linewidth}{!}{
		\begin{tabular}{@{\,}l|ccccc|cccc@{\,}}
& \textsc{bicycle} & \textsc{flowers} & \textsc{garden} & \textsc{stump} & \textsc{treehill} & \textsc{room} & \textsc{counter} & \textsc{kitchen} & \textsc{bonsai} \\
\hline
\BakedSDF{}                         &  \cellcolor{tabfirst}485.4 &  \cellcolor{tabfirst}367.6 &  \cellcolor{tabfirst}384.6 &  \cellcolor{tabfirst}552.5 &  \cellcolor{tabfirst}571.4 &  \cellcolor{tabfirst}534.8 &  \cellcolor{tabfirst}476.2 &  \cellcolor{tabfirst}892.9 &  \cellcolor{tabfirst}584.8 \\
\InstantNGP{}                       &                      9.1 &                      8.8 &                      7.6 &                      13.0 &                      8.2 &                      15.1 &                      8.3 &                      6.4 &                      6.8 \\
\GSplat{}                           &                      177.3 & \cellcolor{tabsecond}342.5 &                      288.2 &                      218.3 & \cellcolor{tabsecond}401.6 &                      168.9 & \cellcolor{tabsecond}321.5 & \cellcolor{tabsecond}311.5 & \cellcolor{tabsecond}303.0 \\
\MERF{} (published)                 &                      195.7 &                      190.5 &                      202.4 &                      147.5 &                      101.7 &                      228.8 &                      183.8 &                      155.3 &                      137.2 \\
\MERF{} (ours)                      &  \cellcolor{tabthird}311.5 &                      274.7 & \cellcolor{tabsecond}343.6 &  \cellcolor{tabthird}264.6 &                      243.3 &  \cellcolor{tabthird}373.1 &  \cellcolor{tabthird}307.7 &                      255.1 &                      206.6 \\
\MERFDistill{}                      & \cellcolor{tabsecond}330.0 &  \cellcolor{tabthird}292.4 &  \cellcolor{tabthird}334.4 & \cellcolor{tabsecond}346.0 &                      211.0 & \cellcolor{tabsecond}390.6 &                      280.1 &  \cellcolor{tabthird}268.1 &  \cellcolor{tabthird}218.3 \\
\MERFDistillTrain{}                 &                      253.2 &                      224.2 &                      223.7 &                      258.4 &                      243.9 &                      336.7 &                      244.5 &                      186.2 &                      148.4 \\
\Ours{} ($\SubmodelResolution = 1$) &                      228.3 &                      213.2 &                      210.1 &                      259.1 &  \cellcolor{tabthird}246.9 &                      330.0 &                      238.1 &                      180.8 &                      141.0 \\
\hline
\ZipNeRF{}                          &                   0.25*       &                    0.25*       &                    0.25*       &                    0.25*       &                    0.25*       &                    0.25*       &                    0.25*       &                    0.25*       &                    0.25*     
		\end{tabular}
	}
	\label{tab:scenes-mipnerf360-fps}
\end{table*}

\let\baselinestretch\savedbaselinestretch
\end{document}